%% file: main.tex
\documentclass[11pt,letterpaper]{article}
\usepackage{rotation-report}

\input{math_commands.tex}

\usepackage{hyperref}
\usepackage{url}
\usepackage{booktabs}
\usepackage{tabularx}
\usepackage{makecell}
\usepackage{graphicx}
\usepackage{multirow}
\usepackage{wrapfig}
\usepackage{adjustbox}
\usepackage[table]{xcolor}
\usepackage{pgf}

\usepackage{array}
\usepackage[skins,breakable]{tcolorbox}
\usepackage{amssymb}

\newtcolorbox{taskbox}[1]{
    enhanced,
    breakable,
    colback=white,
    colframe=ReportLine,
    colbacktitle=ReportPanel,
    coltitle=ReportInk,
    boxrule=0.5pt,
    titlerule=0.5pt,
    arc=1mm,
    left=2.5mm,
    right=2.5mm,
    top=2mm,
    bottom=2mm,
    title={#1},
    fonttitle=\sffamily\bfseries\small,
    before skip=6pt,
    after skip=8pt
}

\newcommand{\taskimage}[3][0.22\linewidth]{%
    \begin{minipage}[t]{#1}
        \centering
        \includegraphics[
            width=\linewidth,
            height=28mm,
            keepaspectratio
        ]{#2}
        \par\vspace{0pt}
        {\scriptsize #3}
    \end{minipage}%
}

\definecolor{RotationCuegain}{HTML}{9AB4CA}
\definecolor{RotationCuedrop}{HTML}{C97481}
\definecolor{modelgroupcolor}{HTML}{B6A7CB}
\colorlet{canonlight}{RotationCuegain!25}
\colorlet{canonmid}{RotationCuegain!50}
\colorlet{canondark}{RotationCuegain!75}

\newcommand{\scoregain}[2]{\makebox[2.8em][r]{#1}\hspace{0.08em}\raisebox{-0.2ex}{\scriptsize\color{black!68}\makebox[3.7em][r]{(#2)}}}
\newcommand{\dc}[3]{\cellcolor{RotationCuegain!#1}\scoregain{#2}{#3}}
\newcommand{\mc}[2]{\cellcolor{RotationCuegain!#1}\makebox[3.4em][r]{#2}}
\newcommand{\dcs}[2]{\scoregain{#1}{#2}}
\newcommand{\grouprule}{\arrayrulecolor{black!30}\specialrule{0.3pt}{0pt}{0pt}\arrayrulecolor{black}}

\newcommand{\plainformat}[1]{#1}
\newcommand{\heatcellbase}[3]{%
  \begingroup
  \pgfmathsetmacro{\score}{#2}\pgfmathsetmacro{\chance}{#3}\pgfmathsetmacro{\diff}{\score-\chance}%
  \ifdim\diff pt=0pt\relax
    \cellcolor{gray!3}#1{#2}%
  \else
    \ifdim\diff pt>0pt\relax
      \def\heatbase{RotationCuegain}\pgfmathsetmacro{\reldiff}{(\score-\chance)/max(100-\chance,1)}%
    \else
      \def\heatbase{RotationCuedrop}\pgfmathsetmacro{\reldiff}{(\chance-\score)/max(\chance,1)}%
    \fi
    \pgfmathtruncatemacro{\shade}{min(68,max(8,8+60*pow(\reldiff,0.65)))}%
    \edef\heatcolor{\heatbase!\shade}\expandafter\cellcolor\expandafter{\heatcolor}#1{#2}%
  \fi
  \endgroup
}
\newcommand{\heatcell}[2]{\heatcellbase{\plainformat}{#1}{#2}}
\newcommand{\bestheatcell}[2]{\heatcellbase{\textbf}{#1}{#2}}
\newcommand{\secondheatcell}[2]{\heatcellbase{\underline}{#1}{#2}}
\newcommand{\secondbest}[1]{\underline{#1}}
\newcommand{\groupdivider}{\arrayrulecolor{black!20}\specialrule{0.5pt}{0pt}{0pt}\arrayrulecolor{black}}

\usepackage{array,siunitx}
\colorlet{tcband}{RotationCuegain!16}
\newcolumntype{N}{S[table-format=2.1,table-number-alignment=center]}

\title{Rotated, but How Far? Diagnosing and Improving Object-Rotation Reasoning in VLMs}

\author{%
Zhaochen Wang\textsuperscript{1}\quad
Yujun Cai\textsuperscript{1}\quad
Huangbo Zou\textsuperscript{2}\quad
Hower Yang\\
Naipeng Dong\textsuperscript{1}\quad
Miao Xu\textsuperscript{1}\quad
Haibin Ling\textsuperscript{2,*}}
\reportaffiliations{\textsuperscript{1}The University of Queensland\qquad
\textsuperscript{2}Westlake University}
\reportcorrespondence{\textsuperscript{*}Corresponding author}
\reportemail{zhaochen.wang@uq.edu.au}
\hypersetup{pdftitle={Rotated, but How Far? Diagnosing and Improving Object-Rotation Reasoning in VLMs},
  pdfauthor={Zhaochen Wang, Yujun Cai, Huangbo Zou, Hower Yang, Naipeng Dong, Miao Xu, Haibin Ling}}
\reportshorttitle{Rotated, but How Far? Diagnosing and Improving Object-Rotation Reasoning in VLMs} % running head
\begin{document}

\maketitle

\begin{abstract}
Vision-language models (VLMs) can detect that an object has rotated across views, but cannot reliably tell by how much. We introduce OR-Bench, a fine-grained benchmark for object-rotation reasoning with eight tasks covering rotation detection, rotation magnitude estimation, and multi-view rotation reasoning. Across 12 VLMs, the gap is stark: the strongest models approach 100\% accuracy on detection, yet even coarse magnitude estimation is near chance. When asked for exact angles, models place 91.8--100\% of their predictions on just 0°, 90°, and 180°, a failure we term \emph{canonical-angle collapse}. This collapse persists even without visual input. Representation probing shows that missing information is only part of the explanation. Although rotation information becomes less recoverable at finer granularity, substantial coarse-grained information remains, and a simple linear probe outperforms the models' generated answers. This suggests that VLMs underuse rotation information they already encode. We therefore propose RotationCue, a lightweight decoder that recovers coarse rotation information from the VLM's own frozen representations and feeds it back to the model as intermediate textual context. Across three VLMs, RotationCue improves every model--task combination on OR-Bench, raising macro-average accuracy by 7.9--12.6 points while preserving general capabilities.
\end{abstract}

\section{Introduction}

Vision-language models (VLMs) \citep{zhang2024vision} are moving from static image understanding toward perception and reasoning in dynamic environments \citep{liu2023visual,chen2024spatialvlm,tong2024cambrian}, where a model must not only recognize what an object is, but also understand how it has changed between observations. Object rotation\footnote{\emph{Object rotation} here means an object's azimuthal rotation about its own vertical axis, with the camera fixed; it is not in-plane image rotation. Section~\ref{sec:OR-Bench} gives the formal definition.} is one such basic and common change. A rotated object can look completely different from one view to the next, yet it is the same object. This makes object rotation a natural test of spatial reasoning \citep{shepard1971mental}. Humans handle this readily: given two views of the same object, people can typically tell whether it has been rotated, roughly estimate by how much, and distinguish smaller rotations from larger ones. Can VLMs do the same?

\begin{figure}
    \centering
    \includegraphics[width=1\linewidth]{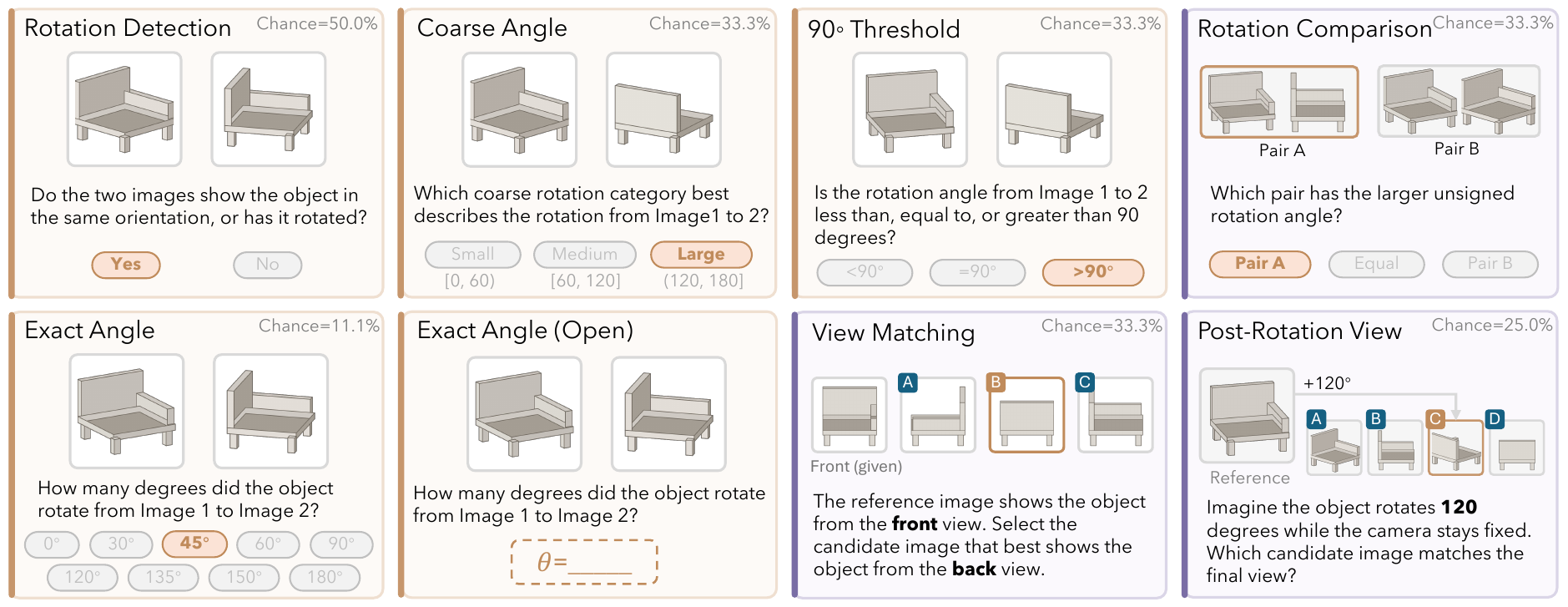}
    \caption{\textbf{Overview of OR-Bench.} Eight tasks span two-view rotation understanding (orange) and multi-view rotation use (purple), covering multiple levels of answer granularity, from detecting whether a rotation occurred to specifying its exact angle.}
    \label{fig:OR-Bench}
\end{figure}

Despite its importance, object rotation remains underexplored in VLM research. Recent benchmarks \citep{zhang2026spinbench,kong2025lrrbench} include rotation-related tasks and consistently identify rotation as a weakness of VLMs. Yet they tell us mainly that models fail, not how: rotation is typically treated as a subtask within a broader benchmark, with coarse evaluation and little analysis of where and why errors arise. We introduce the Object Rotation Benchmark (\textbf{OR-Bench}) to study object rotation systematically, with eight tasks organized into two groups (Figure~\ref{fig:OR-Bench}). Five two-view tasks ask how precisely a model understands the rotation between two observations, from detecting whether it occurred to estimating its coarse angular range and exact angle. Three multi-view tasks ask how well it uses rotation relations across observations: matching views to a reference, comparing rotation magnitudes, and predicting rotated views. Together, these tasks examine how precisely VLMs estimate rotation and how effectively they use it.

\begin{figure}
    \centering
    \includegraphics[width=1\linewidth]{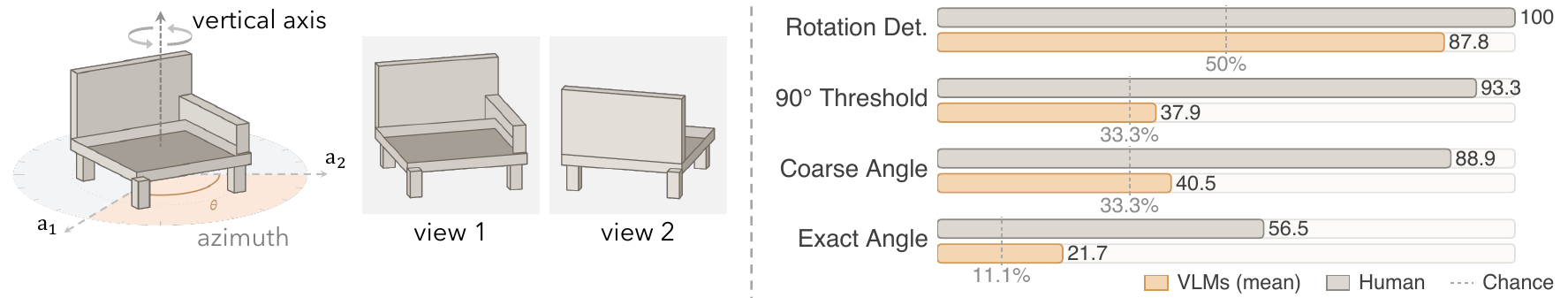}
    \caption{Object-rotation reasoning in OR-Bench. Left: two views of the same object under azimuth rotation. Right: VLMs detect rotation reliably, but performance drops sharply once rotation magnitude must be estimated, even coarsely, leaving a large gap to humans.}
    \label{fig:task_review}
\end{figure}

Our evaluation reveals a sharp boundary (Figure~\ref{fig:task_review}): VLMs can tell \emph{whether} an object has rotated, but not \emph{how far}. Rotation Detection averages 87.8\% across 12 models, with strong models approaching perfect accuracy. Performance drops sharply once models must estimate rotation magnitude, even coarsely: average accuracy falls to 37.9\% on 90° Threshold (comparing the rotation angle with \(90^\circ\)) and 40.5\% on Coarse Angle (estimating its approximate angular range), barely above the 33.3\% chance level and far below human accuracy (93.3\% and 88.9\%, respectively). The failure becomes even more pronounced on the Exact Angle tasks (selecting or directly predicting the exact angle): across models, 91.8–100\% of predictions concentrate on just three canonical angles (0°, 90°, and 180°), a failure mode we term \textbf{canonical-angle collapse}. This collapse persists even when the visual input is removed entirely. 

To understand these failures, we probe the models' internal representations and identify two limitations. First, as angular granularity becomes finer, the models' sensitivity to rotation progressively declines, and less rotation information can be recovered from representations. Second, even where the information is encoded, the models do not fully exploit it: on Coarse Angle in particular, a simple linear probe on the same representations substantially outperforms the models' own predictions. 

If rotation information is encoded but underused, can we put it to work? We propose \textbf{RotationCue}, which takes this information from recoverable to usable. Using external multi-view data, it trains a lightweight decoder to extract coarse-grained, structured rotation cues from a frozen VLM's visual representations. These cues are then fed back to the VLM as intermediate context to support its reasoning. Across three VLMs, RotationCue improves accuracy on all eight OR-Bench tasks, raising the macro-average (Macro-8) by 7.9--12.6 points. With RotationCue, Qwen3-VL-8B outperforms strong proprietary models and a supervised fine-tuning baseline, at modest training cost and without sacrificing general capabilities. Our contributions are threefold:

\begin{itemize}
\item We present OR-Bench, a fine-grained benchmark for object-rotation reasoning that reveals systematic failures in rotation estimation, exhibiting canonical-angle collapse.
\item We trace these failures to two limitations through representation probing: rotation information is less recoverable at finer angular granularity, and where it is recoverable, VLMs fail to fully exploit it.
\item We propose RotationCue, which recovers structured rotation information from frozen representations and reinjects it as intermediate context, improving object-rotation reasoning across tasks and VLMs.
\end{itemize}

\section{Related Work}
\paragraph{Spatial reasoning in VLMs.}
Spatial reasoning benchmarks for VLMs have progressed from pairwise relations in single images \citep{kamath2023whatsup,tong2024cambrian} to spatial understanding \citep{ma2025threedsrbench,wang2025spatial457} and multi-view, video, and dynamic settings \citep{yang2025thinking,zhou2025vlm4d,jia2026omnispatial}, covering spatial relations such as relative position, distance, egocentric direction, and perspective taking. Most closely related, SpinBench \citep{zhang2026spinbench} decomposes perspective taking into fine-grained diagnostic categories, including rotation, and reveals poor rotational understanding across a wide range of VLMs. LRR-Bench \citep{kong2025lrrbench} evaluates rotation alongside spatial movement on synthetic data and likewise identifies it as a persistent weakness. Building on these findings, OR-Bench treats object rotation as a research problem in its own right, evaluating rotation reasoning across controlled angular granularities. Beyond accuracy, we conduct in-depth error and representation analyses.

\paragraph{Probing VLM representations.}
Linear probing \citep{belinkov2022probing} is a standard tool for analyzing internal model representations. Prior work has used it to study depth and surface geometry in vision foundation models \citep{Danier_2025_CVPR}, as well as visual and spatial information in VLMs, such as chart coordinates, object positions, and orientations \citep{fu2025hidden,tartaglini2025diagnosing,kang2026linearmechanismsspatiotemporalreasoning,theodoridis2026probing}. We also adopt linear probing, but go beyond asking whether rotation information is present: we examine what kinds of rotation information are encoded, from coarse to fine, and how well each can be recovered. Building on this analysis, we propose RotationCue, which extracts rotation information from the VLM's own frozen representations and feeds it back as intermediate context. Unlike existing intervention methods \citep{liu2026seeing,li2023inference}, RotationCue neither alters the visual input nor directly edits internal activations, improving rotation reasoning while preserving general-task performance.

\section{OR-Bench: Evaluating Object-Level Rotation}
\label{sec:OR-Bench}

\textbf{Problem Definition.}
OR-Bench studies object rotation about the vertical axis (a change in azimuth), rather than an in-plane rotation of the image. Most objects rest on a supporting surface, and such rotation is the most common in everyday observation. Fixing this axis gives a controlled setting in which to evaluate the object rotation reasoning of VLMs. Given azimuths \(a_1\) and \(a_2\) of two views of the same object, the rotation angle is
\[
\theta=\min\left(\lvert a_1-a_2\rvert,\ 360^\circ-\lvert a_1-a_2\rvert\right)\in[0^\circ,180^\circ].
\]
The angle is unsigned; when a task involves rotation direction, it is provided explicitly in the text.

\textbf{Data Construction.}
OR-Bench uses the \(360^\circ\) turntable subset of the Amazon Berkeley Objects (ABO) \citep{collins2022abo}, with 72 views per product scan at \(5^\circ\) intervals. Frame offsets give the exact angle, so every task carries a noise-free label. From an initial pool of 100 objects, we remove those whose orientation is visually ambiguous due to symmetry and retain 73, from which we build 2{,}109 test examples around nine reference angles: $\{0^\circ,30^\circ,45^\circ,60^\circ,90^\circ,120^\circ,135^\circ,150^\circ,180^\circ\}$. Closed-set tasks balance target classes and answer positions; image-selection candidates use different views of the same object, so the choice depends on rotation rather than object identity.

\textbf{Two-view tasks.}
Five tasks reason over image pairs and assess rotation understanding at different output granularities. \textit{Rotation Detection} determines whether a rotation occurs; \textit{Coarse Angle} classifies the rotation as small (\([0^\circ, 60^\circ)\)), medium (\([60^\circ, 120^\circ]\)), or large (\((120^\circ, 180^\circ]\)); \textit{\(90^\circ\) Threshold} distinguishes rotations below, equal to, or above \(90^\circ\); and \textit{Exact Angle (MCQ)} and \textit{Exact Angle (Open)} require selecting or directly predicting the exact angle, respectively. Coarse Angle and the two Exact Angle variants use the same image pairs, enabling direct comparison across output granularities. 

\textbf{Multi-view tasks.} Three tasks use rotation relations across multiple observations. \textit{View Matching} identifies a specified view (e.g., front, back, left, or right) from candidate images; \textit{Rotation Comparison} compares the magnitudes of two image pairs; and \textit{Post-Rotation View} predicts the resulting view after a text-specified rotation. Overall, the two-view tasks measure how precisely a model captures rotation, from detecting a change to specifying its exact magnitude; the multi-view tasks test how well it matches, compares, and applies rotation relations across observations.

\textbf{Evaluation Protocol.}
We evaluate nine open-source and three proprietary VLMs zero-shot under fixed
input/output formats, counting unparseable responses as incorrect. We report
per-task accuracy, Macro-8 as the unweighted mean across tasks, and Micro-8 over
all examples. Exact Angle (Open) uses normalized exact match, and human
performance provides a reference point. Additional details on benchmark construction, task specifications, evaluation protocols, and random baselines are provided in Appendix~\ref{app:benchmark_details} \& \ref{app:task_examples}.

\begin{table}[t]
\centering
\caption{Results (\%) on OR-Bench. Exact Angle (Open) uses exact-match accuracy. \textbf{Bold} and \underline{underlined} values indicate the best and second-best results within each group. Blue and red indicate above- and below-chance performance, respectively; darker shades indicate larger deviations.}
\label{tab:ObjRot_bench_results}
\small
\setlength{\tabcolsep}{2.0pt}
\renewcommand{\arraystretch}{1.15}
\begin{adjustbox}{max width=\linewidth}
\begin{tabular}{@{}l!{\color{black!30}\vrule width 0.35pt}ccccc!{\color{black!30}\vrule width 0.35pt}ccc!{\color{black!30}\vrule width 0.35pt}cc@{}}
\toprule
\textbf{Model}
& \makecell{\textbf{Rotation}\\\textbf{Detection}}
& \makecell{\(\mathbf{90^\circ}\)\\\textbf{Thresh.}}
& \makecell{\textbf{Coarse}\\\textbf{Angle}}
& \makecell{\textbf{Exact Angle}\\\textbf{(MCQ)}}
& \makecell{\textbf{Exact Angle}\\\textbf{(Open)}}
& \makecell{\textbf{View}\\\textbf{Matching}}
& \makecell{\textbf{Rotation}\\\textbf{Comparison}}
& \makecell{\textbf{Post-Rot.}\\\textbf{View}}
& \makecell{\textbf{Micro}\\\textbf{-8}}
& \makecell{\textbf{Macro}\\\textbf{-8}}
% & \makecell{\textbf{Micro-8}}
% & \makecell{\textbf{Macro-8}} 
\\

\groupdivider
%\rowcolor{blue!8}
\multicolumn{11}{@{}c@{}}{\textit{Proprietary Multimodal Large Language Models}} \\
\groupdivider

GPT-5.4 \citep{openai2026gpt54}
& \secondheatcell{97.9}{50.0}
& \secondheatcell{40.3}{33.3}
& \bestheatcell{53.1}{33.3}
& \bestheatcell{34.6}{11.1}
& \bestheatcell{39.5}{11.1}
& \bestheatcell{75.9}{33.3}
& \bestheatcell{56.4}{33.3}
& \bestheatcell{37.5}{25.0}
& \textbf{59.6}
& \textbf{54.4} \\

GPT-4o \citep{openai2024gpt4o}
& \bestheatcell{99.2}{50.0}
& \heatcell{37.9}{33.3}
& \secondheatcell{42.0}{33.3}
& \secondheatcell{27.2}{11.1}
& \secondheatcell{26.3}{11.1}
& \secondheatcell{68.5}{33.3}
& \secondheatcell{49.4}{33.3}
& \secondheatcell{36.5}{25.0}
& \secondbest{54.2}
& \secondbest{48.4} \\

GPT-4o-mini \citep{openai2024gpt4omini}
& \secondheatcell{97.9}{50.0}
& \bestheatcell{42.0}{33.3}
& \heatcell{38.3}{33.3}
& \heatcell{22.2}{11.1}
& \secondheatcell{26.3}{11.1}
& \heatcell{59.3}{33.3}
& \heatcell{40.3}{33.3}
& \heatcell{27.1}{25.0}
& 50.6
& 44.2 \\

% Open-source group
\groupdivider
%\rowcolor{blue!8}
\multicolumn{11}{@{}c@{}}{\textit{Open-source Multimodal Large Language Models}} \\
\groupdivider

Qwen3-VL-8B \citep{bai2025qwen3vl}
& \bestheatcell{99.6}{50.0}
& \heatcell{37.0}{33.3}
& \heatcell{40.7}{33.3}
& \bestheatcell{24.7}{11.1}
& \secondheatcell{25.5}{11.1}
& \bestheatcell{65.3}{33.3}
& \bestheatcell{48.6}{33.3}
& \secondheatcell{32.8}{25.0}
& \textbf{53.0}
& \textbf{46.8} \\

InternVL3-8B \citep{zhu2025internvl3}
& \heatcell{96.5}{50.0}
& \secondheatcell{39.1}{33.3}
& \secondheatcell{42.0}{33.3}
& \secondheatcell{22.2}{11.1}
& \bestheatcell{28.4}{11.1}
& \heatcell{47.2}{33.3}
& \heatcell{39.5}{33.3}
& \heatcell{29.2}{25.0}
& 49.5
& \secondbest{43.0} \\

InternVL3.5-8B \citep{wang2025internvl35}
& \secondheatcell{99.2}{50.0}
& \heatcell{37.4}{33.3}
& \heatcell{32.1}{33.3}
& \secondheatcell{22.2}{11.1}
& \heatcell{23.5}{11.1}
& \heatcell{49.1}{33.3}
& \secondheatcell{44.0}{33.3}
& \bestheatcell{36.5}{25.0}
& \secondbest{49.5}
& 43.0 \\

Qwen2.5-VL-7B \citep{bai2025qwen25vl}
& \heatcell{86.0}{50.0}
& \heatcell{37.4}{33.3}
& \heatcell{41.6}{33.3}
& \secondheatcell{22.2}{11.1}
& \heatcell{23.0}{11.1}
& \secondheatcell{56.0}{33.3}
& \heatcell{41.6}{33.3}
& \heatcell{30.2}{25.0}
& 47.4
& 42.3 \\

InternVL3.5-4B \citep{wang2025internvl35}
& \heatcell{97.7}{50.0}
& \bestheatcell{41.2}{33.3}
& \heatcell{38.3}{33.3}
& \heatcell{20.6}{11.1}
& \heatcell{22.6}{11.1}
& \heatcell{48.6}{33.3}
& \heatcell{42.0}{33.3}
& \heatcell{26.0}{25.0}
& 48.8
& 42.1 \\

Qwen2-VL-7B \citep{wang2024qwen2vl}
& \heatcell{62.8}{50.0}
& \heatcell{38.3}{33.3}
& \bestheatcell{45.7}{33.3}
& \heatcell{16.0}{11.1}
& \heatcell{20.2}{11.1}
& \heatcell{54.6}{33.3}
& \heatcell{34.2}{33.3}
& \heatcell{27.6}{25.0}
& 40.4
& 37.4 \\

Cosmos-Reason1-7B \citep{azzolini2025cosmos}
& \heatcell{65.0}{50.0}
& \heatcell{36.2}{33.3}
& \heatcell{38.3}{33.3}
& \heatcell{21.0}{11.1}
& \heatcell{22.2}{11.1}
& \heatcell{52.8}{33.3}
& \heatcell{36.6}{33.3}
& \heatcell{25.0}{25.0}
& 40.4
& 37.1 \\

Phi-3.5-Vision \citep{abdin2024phi3}
& \heatcell{90.7}{50.0}
& \heatcell{33.7}{33.3}
& \heatcell{37.9}{33.3}
& \heatcell{16.5}{11.1}
& \heatcell{12.3}{11.1}
& \heatcell{38.4}{33.3}
& \heatcell{32.9}{33.3}
& \heatcell{24.0}{25.0}
& 42.4
& 35.8 \\

LLaVA-OneVision-7B \citep{li2024llavaonevision}
& \heatcell{60.5}{50.0}
& \heatcell{33.7}{33.3}
& \heatcell{35.8}{33.3}
& \heatcell{11.1}{11.1}
& \heatcell{14.0}{11.1}
& \heatcell{37.0}{33.3}
& \heatcell{35.0}{33.3}
& \heatcell{25.5}{25.0}
& 35.0
& 31.6 \\

\midrule
\rowcolor{black!5}
\textbf{Human}
& 100.0
& 93.3
& 88.9
& 56.5
& 56.5
& 100.0
& 72.1
& 87.2
& --
& 81.8 \\

\bottomrule
\end{tabular}
\end{adjustbox}
\end{table}

\subsection{Overall Performance and Trends}
Current VLMs show a sharp boundary in rotation understanding: they can reliably tell whether an object has rotated, but struggle to judge how far (Table~\ref{tab:ObjRot_bench_results}). Across the nine open-source models, Rotation Detection averages 84.2\%, with the strongest models approaching ceiling accuracy. Once rotation magnitude is required, however, performance falls close to chance: \(90^\circ\) Threshold and Coarse Angle average only 37.1\% and 39.2\%, respectively, against a 33.3\% chance level, far below human accuracy (93.3\% and 88.9\%).

A similar pattern appears in the multi-image tasks. View Matching, which only requires identifying a named view such as front or left, averages 49.9\%. Performance drops when models must compare rotation magnitudes (Rotation Comparison, 39.4\%) and drops further when they must apply a rotation to predict a new view (Post-Rotation View, 28.5\%). Proprietary models follow the same pattern at higher absolute performance. Across both two-view and multi-view tasks, models are stronger at detecting or matching rotations than at estimating or using rotation magnitude.

\begin{figure*}
\centering
\newlength{\tblpanelw}
\setlength{\tblpanelw}{168pt}
\begin{minipage}[c]{\tblpanelw}
\centering
\small
\setlength{\tabcolsep}{2.0pt}
\renewcommand{\arraystretch}{1.15}
\begin{tabular}{@{}l!{\color{black!30}\vrule width 0.35pt}ccc@{}}
\toprule
& \multicolumn{3}{c}{Exact Angle Task} \\
\cmidrule(lr){2-4}
\textbf{Model} & \textbf{MCQ} & \textbf{Open} & \makecell{\textbf{MCQ}\\[-1pt]\scriptsize\textbf{(text-only)}} \\
\midrule
\textbf{Qwen3-VL-8B} & \cellcolor{canonlight}91.8 & \cellcolor{canonmid}97.1 & \cellcolor{canonmid}97.5 \\
InternVL3.5-8B & \cellcolor{canonmid}98.4 & \cellcolor{canondark}100.0 & \cellcolor{canondark}100.0 \\
Qwen2.5-VL-7B & \cellcolor{canondark}99.6 & \cellcolor{canondark}100.0 & \cellcolor{canondark}100.0 \\
\bottomrule
\end{tabular}
\end{minipage}%
\hfill
\begin{minipage}[c]{\dimexpr\textwidth-\tblpanelw-12pt\relax}
\centering
\includegraphics[width=\linewidth]{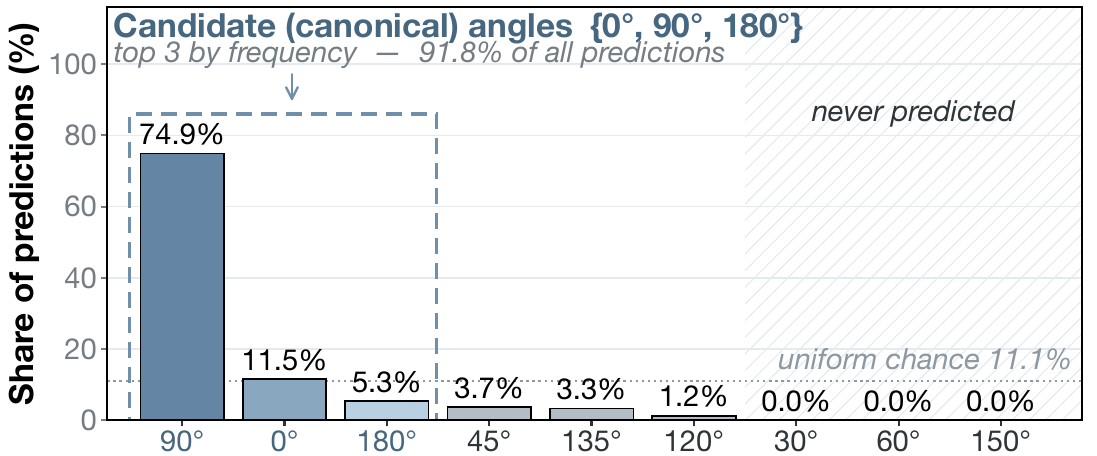}
\end{minipage}
\caption{Canonical-angle collapse across input and response settings. \textbf{Left:} share of predictions in the canonical set $\{0^\circ,90^\circ,180^\circ\}$; darker cells indicate stronger collapse. \textbf{Right:} Qwen3-VL-8B's Exact Angle (MCQ) predictions concentrate primarily at $90^\circ$.}
\label{fig:canonical_angle_collapse}
\end{figure*}

\subsection{Canonical-Angle Collapse}
\label{sec:Canonical-Angle}
Beyond accuracy, human errors still reflect a coherent estimate of object rotation. Although human performance on Exact Angle is also limited (56.5\%), their errors are typically local, such as confusing \(30^\circ\) with \(45^\circ\), while remaining consistent across granularities. VLM errors do not show this consistency. We closely examine three representative open-source models and find that they concentrate 91.8\%, 98.4\%, and 99.6\% of their predictions on just three canonical angles (\(0^\circ/90^\circ/180^\circ\)). We call this behavior \textbf{canonical-angle collapse}, with \(90^\circ\) emerging as an especially dominant, often default, response (Figure~\ref{fig:canonical_angle_collapse}). Closed-source models show a similar pattern (Appendix~\ref{sec:appendix_proprietary_collapse}).

The collapse persists even without visual input: after removing all visual tokens, 97.5--100\% of predictions still fall on the three canonical angles, indicating a strong textual prior toward these values. Visual input still matters, however: it changes 37\% of predictions and raises accuracy from the 11.1\% chance level to 22.2--24.7\%. Yet this gain is concentrated almost entirely on \(0^\circ\), which reaches 100\% recall and 89.0\% precision. By contrast, \(90^\circ\) accounts for 67--75\% of all predictions but attains only 12.9\% precision, barely above its 11.1\% base rate, while accuracy on noncanonical angles remains at 0--2.5\%. This pattern cannot be explained simply by rounding to the nearest canonical angle, which would predict \(0^\circ\) for \(30^\circ\) or \(180^\circ\) for \(150^\circ\). Instead, the models can reliably separate zero from nonzero rotation, but once the rotation is nonzero, their exact-angle predictions carry little information about its magnitude.

\section{Explaining the Rotation Reasoning Boundary}
\label{sec:linear_probe}
Why does rotation reasoning deteriorate once finer-grained rotation information
is required? A natural explanation is that VLMs do not capture rotation magnitude
from the visual input at all. To test this, we examine whether the relevant rotation information is present in the internal representations of three representative models. We focus on three increasingly fine-grained two-view tasks: Rotation Detection, Coarse Angle, and Exact Angle. We train linear RidgeClassifier probes \citep{hoerl1970ridge} on frozen vision encoder (VE) and LLM representations using five-fold object-disjoint cross-validation with out-of-fold (OOF) predictions \citep{belinkov2022probing}. For comparison, we evaluate the model's own answer, i.e., its zero-shot prediction, on the same examples. Both are scored with reference-adjusted balanced accuracy (AdjBA; 0 = chance, 100 = perfect) \citep{brodersen2010balanced}. Full details are provided in Appendix~\ref{sec:appendix_probe}.

\subsection{Probing the Vision Encoder}
\label{sec:VE_probe}

We first probe the vision encoder to test whether it captures rotation-magnitude
information. Since the VE encodes the two images independently, we use the simplest absolute difference (\texttt{absdiff}), \textit{i.e.}, $\lvert f_1 - f_2 \rvert$ on their frozen representations \(f_1\) and \(f_2\). %:
This operation is fixed and parameter-free: it adds no learnable relational module, so any rotation information recovered here is already carried by the two single-image representations. 

\begin{figure}[!t]
    \centering
    \includegraphics[width=1\linewidth]{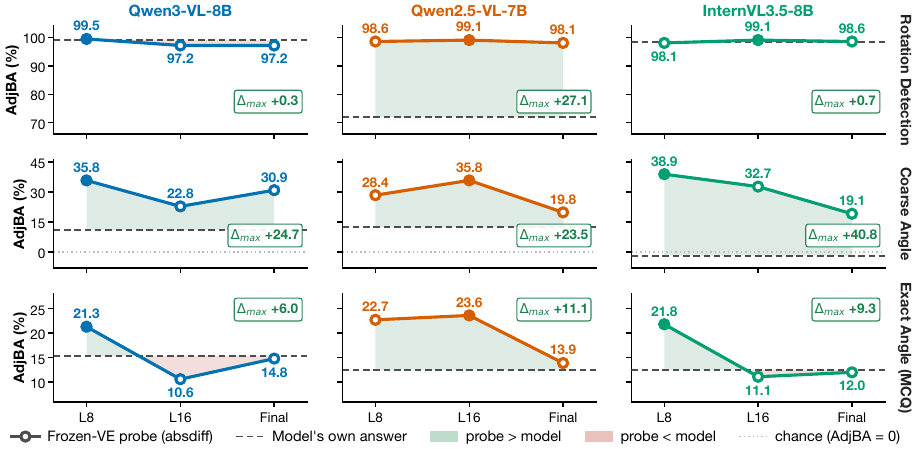}
    \caption{Linear probe accuracy at L8, L16, and the final VE layer, with the
model's own answer as a reference. Recoverability declines from
Rotation Detection to Coarse Angle to Exact Angle.}
    \label{fig:VE_probe}
\end{figure}

We probe L8, L16, and the final VE layer. Across all three models, rotation information shows a clear coarse-to-fine decline in recoverability (Figure~\ref{fig:VE_probe}). Rotation Detection is nearly perfectly recoverable (98.0--98.6 AdjBA, averaged across layers), Coarse Angle remains substantially recoverable (28.0--30.2), and Exact Angle falls further (15.0--20.1). At many layers the probe is also more accurate than the model's own answer, most consistently for Coarse Angle, where the largest per-layer margin reaches \(+24.7\), \(+23.5\), and \(+40.8\) AdjBA for the three models. As a sanity check, replacing the true Coarse Angle labels with random labels drops probe accuracy from 28.4--38.9 AdjBA to between \(-6.2\) and \(-3.8\) (Appendix~\ref{sec:appendix_random_control}), indicating that the signal reflects genuine rotation information.

\begin{figure}[!t]
    \centering
    \includegraphics[width=1\linewidth]{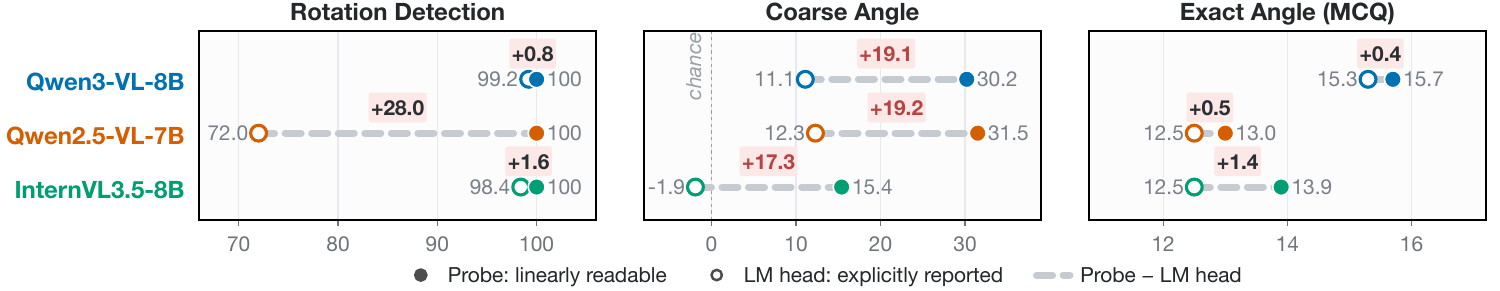}
    \caption{Probe vs.\ LM head at \(h_{\mathrm{pre}}\). Probe (filled) and LM head (open) are two linear heads reading the same pre-answer representation. The gap is largest on Coarse Angle.}
    \label{fig:hpre}
\end{figure}

\subsection{Probing at the Output Head}
\label{sec:hpre_probe}
We next probe \(h_{\mathrm{pre}}\), the final-layer hidden state at the last prompt position, from which the LM head produces the model's own answer. Unlike the VE analysis, we apply no \texttt{absdiff} or other external relational operation. The probe and the LM head are therefore two linear heads reading the same representation, which makes the comparison direct at the representation level.

At \(h_{\mathrm{pre}}\) we observe a similar coarse-to-fine decline as in the VE: across the three models, probes reach 100.0 AdjBA on Rotation Detection, 15.4--31.5 on Coarse Angle, and 13.0--15.7 on Exact Angle. The rotation relation between the two views is therefore already encoded in the model's own representation, and remains linearly readable there. The probe again outperforms the LM head, most clearly on Coarse Angle, where it leads by 17.3--19.2 AdjBA (Figure~\ref{fig:hpre}). Rotation Detection is already near ceiling, and on Exact Angle the two remain close, differing by only
0.4--1.4 AdjBA.

The models' failure at finer granularities thus reflects two complementary limitations: as the angle becomes finer, the recoverable information weakens, and what remains is not fully exploited at readout. On Rotation Detection the information is almost perfectly recoverable and the model generally reports it correctly. On Coarse Angle recoverability drops but stays substantial, yet the gap to the LM head is largest, so the available information is under-used. 
On Exact Angle both the information and the answer are weak: the model reliably resolves only whether the rotation is zero, and elsewhere falls back on its textual prior, consistent with the canonical-angle collapse observed earlier (Section~\ref{sec:Canonical-Angle}). 

This gap is not specific to Coarse Angle task or to how the probes are trained: it also appears on $90^\circ$ Threshold, another coarse-granularity task using different images and prompts, and persists when the probes are trained on an external dataset (Appendix~\ref{sec:appendix_threshold_probe}).

\section{RotationCue: Rotation Context from Frozen Representations}
\begin{figure}[!t]
    \centering
    \includegraphics[width=1\linewidth]{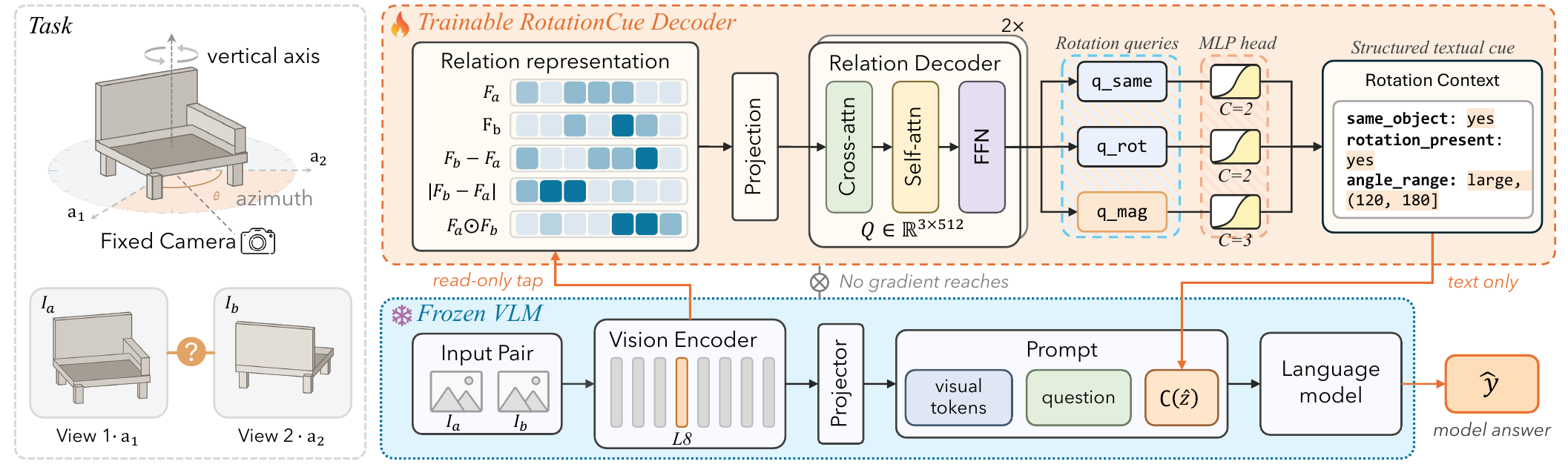}
    \caption{\textbf{RotationCue.} The relation representation is read from vision layer L8 and passed to a lightweight decoder, without modifying the frozen VLM. Three learnable rotation queries predict a three-field rotation context, which is reinjected into the VLM as text only.}
    \label{fig:RCue}
\end{figure}

The probing results show that VLM representations retain substantial rotation
information, particularly at coarse angular granularities. This raises a natural
question: can this information be further utilized to improve object rotation
reasoning? We introduce \textbf{RotationCue}, which uses a lightweight decoder to recover rotation information from internal representations as structured context, and reinjects it into the frozen VLM for reasoning. All rotation information supplied to the model is decoded from its own frozen representations. RotationCue thus extends the probing analysis from diagnosis to improvement.

\subsection{Recovery and Reinjection}

RotationCue reads rotation information directly from the frozen visual
representations, before language generation. Guided by the probing results in
Section~\ref{sec:VE_probe}, we tap L8, where coarse rotation information remains
reliably recoverable across all three models. Given two views \(I_A\) and
\(I_B\), we extract \(F_A=V^{(8)}(I_A)\) and \(F_B=V^{(8)}(I_B)\). To recover
richer rotation information from the two views, we construct the following
pairwise relation representation:
\[
R_{\mathrm{full}}(F_A,F_B)
=
[F_A; F_B; F_B-F_A; \lvert F_A - F_B \rvert; F_A\odot F_B].
\]
\textbf{Structured rotation context.}
RotationCue recovers three rotation-related fields as structured context: \texttt{angle\_range}, \texttt{same\_object}, and \texttt{rotation\_present}. The primary field, \texttt{angle\_range}, represents rotation magnitude as \texttt{small}
\((0^\circ, 60^\circ)\), \texttt{medium} \([60^\circ, 120^\circ]\), or
\texttt{large} \((120^\circ, 180^\circ]\). We choose this granularity because
Coarse Angle retains substantial recoverable information while showing the
largest gap from the model's own predictions. The other two fields are conditions: \texttt{same\_object} determines whether the views depict the same object, and \texttt{rotation\_present} determines whether a rotation occurs. The angle range is provided only when both hold, preventing irrelevant rotation
context from being injected.

\(R_{\mathrm{full}}\) is projected into a low-dimensional space and aggregated by three learnable rotation queries. Three heads predict \texttt{same\_object} \(\rightarrow\) \texttt{rotation\_present} \(\rightarrow\) \texttt{angle\_range}, with an angular-distance loss penalizing larger cross-bin errors. The decoded outputs \(\hat{z}=D_{\phi}(R_{\mathrm{full}})\) are serialized with a fixed template and reinjected as context: \(\hat y=M(I_A,I_B,q,C(\hat z))\). Only the decoder is trained; the base VLM remains frozen. Full architectural, serialization, and training details are provided in the Appendix~\ref{sec:appendix_rotationcue}. Training uses only external multi-view data: ALOI-View \citep{geusebroek2005aloi} and SelectiveMV ModelNet40v1 \citep{alzahrani2024selectivemv,wu2015shapenets}, with object-disjoint train/val splits. OR-Bench is held out entirely for evaluation. The same readout layer, relation representation, decoder design, and training protocol are used across all three models, with one decoder per model serving all tasks.

\subsection{Results on OR-Bench}

\begin{table}
\centering
\caption{Task-level RotationCue results on OR-Bench across three VLMs. Main values are scores (\%); smaller lower-right values in parentheses are gains over Raw ($\Delta$, percentage points), with shading encoding $\Delta$. Mean $\Delta$ averages the gain across the three models.}
\label{tab:RotationCue_task_results}
\begingroup
\small
\setlength{\tabcolsep}{3.5pt}
\renewcommand{\arraystretch}{1.12}
\begin{tabular}{@{}llccc!{\color{black!25}\vrule width 0.3pt}c@{}}
\toprule
Level & Task & Qwen3-VL-8B & Qwen2.5-VL-7B & InternVL3.5-8B & Mean $\Delta$ \\
\midrule

\multirow{5}{*}{Two-view}
& Rotation Detection
& \dc{8}{100.0}{+0.4}
& \dc{38}{99.0}{+13.0}
& \dc{8}{100.0}{+0.8}
& \mc{16}{+4.7} \\

& Coarse Angle
& \dc{62}{66.7}{+26.0}
& \dc{62}{63.4}{+21.8}
& \dc{62}{60.5}{+28.4}
& \mc{62}{+25.4} \\

& $90^\circ$ Threshold
& \dc{50}{55.6}{+18.6}
& \dc{50}{57.2}{+19.8}
& \dc{26}{44.0}{+6.6}
& \mc{50}{+15.0} \\

& Exact Angle (MCQ)
& \dc{26}{32.9}{+8.2}
& \dc{38}{33.7}{+11.5}
& \dc{16}{28.0}{+5.8}
& \mc{26}{+8.5} \\

& Exact Angle (Open)
& \dc{26}{34.6}{+9.1}
& \dc{16}{27.2}{+4.1}
& \dc{16}{27.2}{+3.7}
& \mc{16}{+5.6} \\

\grouprule

\multirow{3}{*}{Multi-view}
& View Matching
& \dc{8}{67.6}{+2.3}
& \dc{8}{57.4}{+1.4}
& \dc{8}{50.9}{+1.8}
& \mc{8}{+1.8} \\

& Rotation Comparison
& \dc{26}{57.2}{+8.6}
& \dc{50}{58.4}{+16.8}
& \dc{38}{57.2}{+13.2}
& \mc{38}{+12.9} \\

& Post-Rotation View
& \dc{50}{49.0}{+16.2}
& \dc{38}{42.7}{+12.5}
& \dc{8}{39.1}{+2.6}
& \mc{38}{+10.4} \\

\midrule
\rowcolor{black!5}
\multicolumn{2}{l}{Macro-8 (all tasks)}
& \dcs{58.0}{+11.2}
& \dcs{54.9}{+12.6}
& \dcs{50.9}{+7.9}
& +10.5 \\

\bottomrule
\end{tabular}
\endgroup
\end{table}

RotationCue improves all 24 model--task combinations across three models and eight tasks, with Macro-8 gains of 11.2, 12.6, and 7.9 percentage points for Qwen3-VL-8B, Qwen2.5-VL-7B, and InternVL3.5-8B, respectively. With RotationCue, Qwen3-VL-8B reaches 58\% Macro-8, surpassing GPT-5.4 at 54.4\%. Full task-level results are reported in Table~\ref{tab:RotationCue_task_results}. The largest gains occur on tasks directly aligned with the recovered rotation cue. Coarse Angle improves by 21.8--28.4 points. Rotation Detection reaches 100.0\% for Qwen3-VL-8B and InternVL3.5-8B, while Qwen2.5-VL-7B improves from 86.0\% to 99.0\%. Importantly, the gains extend beyond tasks directly specified by the cue. Although \texttt{angle\_range} provides only three coarse angular intervals, Exact Angle (MCQ) improves by 5.8--11.5 points, Rotation Comparison by 8.6--16.8 points, and Post-Rotation View by 2.6--16.2 points, with consistent gains also on Exact Angle (Open) and the $90^\circ$ Threshold task.

We attribute these gains to RotationCue acting as an explicit textual
anchor: it turns rotation information from the model's own representations into text, bridging what the model encodes visually and what it can use in language reasoning. RotationCue also substantially mitigates canonical-angle collapse, reducing the canonical-angle prediction rate from 91.8\%--99.6\% to 59.3\%--78.6\%. Taking a closer look at Qwen3-VL-8B on Exact Angle (MCQ), we find that when the cue is correct, accuracy on non-zero rotations rises from 19.4\% to 40.3\% on the same items, and the prediction distribution becomes more plausible.

\subsection{Context Source and Quality}
\label{sec:RotationCue_comparison}

\begin{table}[t]
\centering
\caption{Macro-8 (\%) with (a) different intermediate-context strategies and (b) RotationCue content quality. Self-gen uses the same three-field format as RotationCue; Shuffled uses a cue from another example; Oracle uses ground-truth rotation information.}
\label{tab:RotationCue_context_quality}
\begingroup
\small
\setlength{\tabcolsep}{2.4pt}
\renewcommand{\arraystretch}{1.18}
\begin{tabular}{@{}l NNNN >{\columncolor{tcband}}N @{\hspace{8pt}} NN >{\columncolor{tcband}}N N@{}}
\toprule
& \multicolumn{5}{c}{(a) Source of intermediate context} & \multicolumn{4}{c}{(b) RotationCue content quality} \\
\cmidrule(lr){2-6}\cmidrule(lr){7-10}
{Model} & {Raw} & {ZS-CoT} & {5-shot} & {Self-gen.} & {RotationCue} & {Shuffled} & {Raw} & {RotationCue} & {Oracle} \\
\midrule
Qwen3-VL-8B & 46.8 & 47.1 & 52.3 & 48.6 & \bfseries 58.0 & 44.4 & 46.8 & 58.0 & \bfseries 65.3 \\
Qwen2.5-VL-7B & 42.3 & 43.1 & 44.4 & 47.1 & \bfseries 54.9 & 38.7 & 42.3 & 54.9 & \bfseries 63.6 \\
InternVL3.5-8B & 43.0 & 39.7 & 44.4 & 47.4 & \bfseries 50.9 & 42.9 & 43.0 & 50.9 & \bfseries 59.6 \\
\bottomrule
\end{tabular}
\endgroup
\end{table}

We further compare RotationCue with alternative strategies for supplying
intermediate context (Table~\ref{tab:RotationCue_context_quality}). Besides
CoT-based methods (zero-shot CoT and 5-shot CoT), we prompt the model to generate
the same three-field context used by RotationCue before producing its final
answer (Self-gen), which holds the format fixed and changes only where the
content comes from. RotationCue performs best on all three models, exceeding
5-shot CoT by 5.7, 10.4, and 6.5 Macro-8 points. Free-form reasoning alone brings
little benefit: Zero-shot CoT brings almost no improvement over Raw and even
hurts performance on InternVL3.5-8B, while the more structured Self-gen yields
consistently better results. 

We then keep the three-field structure fixed and vary only the quality of its content. Performance consistently follows
\(\text{Shuffled} < \text{Raw} < \text{RotationCue} < \text{Oracle}\). Shuffled
cues lower Macro-8 by 0.1--3.5 points, while Oracle yields a further
7.4--8.7-point gain over RotationCue. Together, these results show that
RotationCue's advantage comes from both structured, task-specific intermediate
context and more accurate rotation information.

RotationCue is also more efficient and stable than free-form CoT. Under the same inference setup, it adds only 1.2\% latency over Raw while being about
\(2.4\times\) faster than zero-shot CoT. Its intermediate context is compact, and under stochastic sampling it produces more consistent predictions than free-form CoT while maintaining a clear accuracy advantage.

\subsection{Why Not Just Fine-Tune the Base Model?}

\begin{table}
\centering
\caption{Comparison between RotationCue and Rotation-SFT on Qwen3-VL-8B (\%). RotationCue achieves larger gains on OR-Bench while largely preserving performance on external benchmarks. Parenthesized values denote changes from the raw model, computed from displayed scores.}
\label{tab:RotationCue_rotation_sft}
\small
\setlength{\tabcolsep}{4pt}
\renewcommand{\arraystretch}{1.18}
\resizebox{\linewidth}{!}{%
\begin{tabular}{@{}lccccccc@{}}
\toprule
& \multicolumn{2}{c}{\textbf{OR-Bench}} & \multicolumn{5}{c}{\textbf{External benchmarks}} \\
\cmidrule(lr){2-3}\cmidrule(lr){4-8}
\textbf{Method} & \textbf{Macro-8} & \textbf{Micro-8} & \textbf{VSR-ZS} & \textbf{CV 2D} & \textbf{CV 3D} & \textbf{CV Avg.} & \textbf{SpinBench} \\
\midrule
Qwen3-VL-8B (Raw) & 46.8 & 53.0 & 83.6 & 80.1 & 93.3 & 86.7 & 67.0 \\
Rotation-SFT & 51.2 (+4.4) & 56.3 (+3.3) & 75.6 (\(-8.0\)) & 75.4 (\(-4.7\)) & 89.0 (\(-4.3\)) & 82.2 (\(-4.5\)) & 61.8 (\(-5.2\)) \\
\rowcolor{tcband}
\textbf{RotationCue} & \textbf{58.0 (+11.2)} & \textbf{62.9 (+9.9)} & 84.0 (+0.4) & 80.1 (---) & 92.5 (\(-0.8\)) & 86.3 (\(-0.4\)) & \textbf{69.2 (+2.2)} \\
\bottomrule
\end{tabular}%
}
\end{table}

RotationCue keeps the base VLM frozen and places rotation supervision externally, in a lightweight decoder that recovers rotation information from its representations. An alternative is to apply comparable supervision internally, by updating trainable parameters within the VLM itself. Rotation-SFT does this: it uses the same
external multi-view sources and angular semantics, reformulated as visual
question--answer pairs, and fine-tunes the language side through LoRA while
keeping the vision encoder and projector frozen. Training details and per-task results are provided in the Appendix~\ref{sec:appendix_rotation_sft}.

Rotation-SFT underperforms RotationCue on both aggregate metrics, scoring 51.2
versus 58.0 on Macro-8 and 56.3 versus 62.9 on Micro-8, corresponding to gaps of
6.8 and 6.6 points. Its gains are also concentrated: Rotation-SFT performs
strongest on Coarse Angle, the task most directly aligned with its supervision,
while showing no consistent improvement across the remaining OR-Bench tasks and
even degrading on several of them. 

This specialization also affects general performance. Relative to the base model,
Rotation-SFT drops by 8.0 points on VSR-ZS \citep{liu2023visual}, 4.5 on CV-Bench
\citep{tong2024cambrian}, and 5.2 on SpinBench \citep{zhang2026spinbench}.
RotationCue leaves the base model unchanged and largely preserves its performance
on these benchmarks, and requires only 0.23 A100 GPU-hours to train, compared
with roughly 42 A100 GPU-hours for Rotation-SFT.

\section{Conclusion}
Object rotation is a basic form of visual change, and current VLMs handle it in a strikingly uneven way: they reliably register that an object has turned, but not how far. OR-Bench makes this failure precise, showing that it is not a gradual loss of accuracy with finer granularity but a break at the point where a
magnitude must be produced at all. Representation probing then locates it in two
places: recoverable rotation information weakens as granularity becomes finer,
and the readout leaves unused much of what remains. RotationCue puts that unused
information to work, recovering rotation state from frozen representations and
returning it to the model as context. Across three different VLMs, it improves object-rotation reasoning on every OR-Bench task, without updating a VLM weight or compromising general capability.

\bibliography{references}
\bibliographystyle{iclr2027_conference}

\reportappendix

\section{OR-Bench Construction and Composition}
\label{app:benchmark_details}

OR-Bench evaluates object-level rotation reasoning through eight tasks: five two-view tasks assess the same rotation relation at different output granularities, and three multi-view tasks require models to match, compare, or apply rotation relations across observations. The benchmark contains 2{,}109 test examples constructed from 73 objects.

\subsection{Source Data and Rotation Labels}

OR-Bench uses the \(360^\circ\) turntable subset of Amazon Berkeley Objects (ABO) \citep{collins2022abo}. Each product scan contains 72 views captured at \(5^\circ\) azimuthal intervals. The target rotation is the unsigned shortest azimuthal displacement between two views of the same object rotated about its vertical axis, rather than an in-plane image rotation.

For frame indices \(i,j\in\{0,\ldots,71\}\), let \(k=|i-j|\). The relative rotation angle is
\[
\theta = 5^\circ \min(k,72-k) \in [0^\circ,180^\circ].
\]
This is equivalent to the azimuth-based definition in Section~\ref{sec:OR-Bench}. Frame offsets therefore provide exact rotation labels without pose estimation or manual angle annotation.

The target describes the angular separation between endpoint views. When a task requires rotation direction, as in Post-Rotation View, the direction is provided explicitly in the text.

\subsection{Object Screening}

\begin{figure}[!ht]
    \centering
    \includegraphics[width=0.5\linewidth]{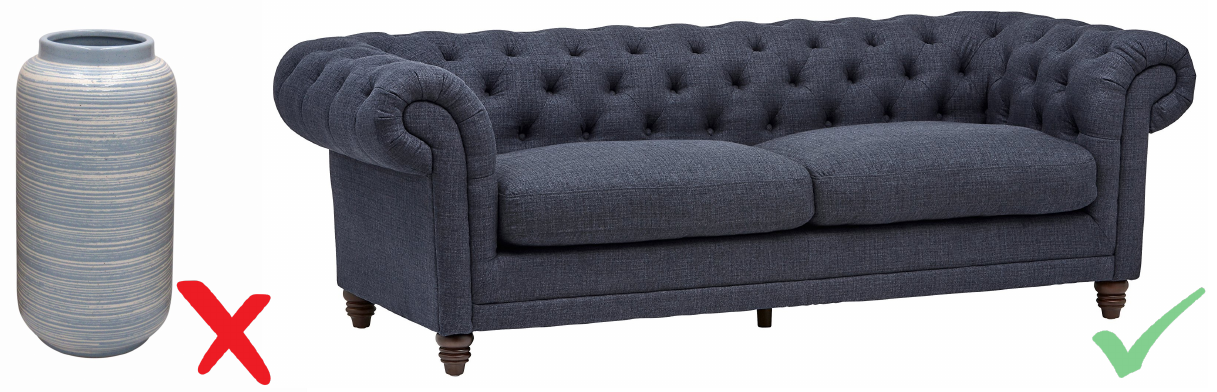}
    \caption{\textbf{Illustrative examples of object screening.} Objects whose orientation is visually ambiguous due to symmetry, such as the vase (left), are excluded. Objects with distinctive orientation cues, such as the sofa (right), are retained.}
    \label{fig:object_screening}
\end{figure}

We remove objects whose orientation is visually ambiguous due to symmetry around the vertical axis. For these objects, different azimuths can produce nearly indistinguishable views despite having different rotation labels. Screening retains objects with distinguishable orientation cues, yielding 73 objects for benchmark construction. Figure~\ref{fig:object_screening} illustrates the screening criterion.

\subsection{Reference Angles and Task Organization}

Tasks are constructed around nine reference angles:
\[
\{0^\circ,30^\circ,45^\circ,60^\circ,90^\circ,120^\circ,135^\circ,150^\circ,180^\circ\}.
\]
These angles provide a symmetric set of representative rotations spanning the full \(0^\circ\)–\(180^\circ\) range, including canonical angles and common intermediate angles.

\textbf{Two-view tasks.} Five tasks query the rotation relation within a single image pair. \textit{Rotation Detection} asks whether the orientation changes. \textit{Coarse Angle} assigns the rotation to one of three intervals: small (\([0^\circ,60^\circ)\)), medium (\([60^\circ,120^\circ]\)), or large (\((120^\circ,180^\circ]\)). \textit{\(90^\circ\) Threshold} distinguishes rotations below, equal to, or above \(90^\circ\). \textit{Exact Angle (MCQ)} requires selecting the exact angle, while \textit{Exact Angle (Open)} requires predicting it directly.

Coarse Angle and the two Exact Angle variants use the same image pairs. Each pair is associated with a coarse interval label and an exact angle label, with the latter queried in both multiple-choice and open-response formats. This shared-input design enables comparison across output granularities and response formats while holding the visual input fixed. 

\textbf{Multi-view tasks.} Three tasks require rotation relations to be used across multiple observations. \textit{View Matching} identifies a specified object view, such as front, back, left, or right, from candidate images. \textit{Rotation Comparison} compares the rotation magnitudes of two image pairs. \textit{Post-Rotation View} selects the resulting view after a text-specified rotation. Together, these tasks assess whether models can use rotation relations for view matching, magnitude comparison, and transformation prediction.

\subsection{Composition and Balancing}

The eight tasks contain 2{,}109 test examples from 73 objects. Because source images can be reused across examples, these examples contain 5{,}712 image occurrences drawn from 2{,}392 distinct source images. Closed-set tasks balance target classes and correct answer positions. For image-selection tasks, all candidate images depict the same object, preventing object identity from serving as a shortcut. Table~\ref{tab:benchmark_composition} summarizes the benchmark composition.

\begin{table}[t]
    \centering
    \caption{\textbf{Composition of OR-Bench.} Uniform reference denotes uniform guessing over the answer choices for closed-set tasks and over the nine reference angles for Exact Angle (Open).}
    \label{tab:benchmark_composition}
    \small
    \setlength{\tabcolsep}{7pt}
    \renewcommand{\arraystretch}{1.08}
    \begin{tabular}{@{}llccc@{}}
        \toprule
        \textbf{Task Group} & \textbf{Task} & \textbf{Examples} & \textbf{Imgs./Ex.} & \textbf{Uniform reference} \\
        \midrule
        \multirow{5}{*}{Two-view}
        & Rotation Detection     & 486 & 2 & 50.0\% \\
        & Coarse Angle           & 243 & 2 & 33.3\% \\
        & \(90^\circ\) Threshold & 243 & 2 & 33.3\% \\
        & Exact Angle (MCQ)      & 243 & 2 & 11.1\% \\
        & Exact Angle (Open)     & 243 & 2 & 11.1\% \\
        \cmidrule(lr){1-5}
        \multirow{3}{*}{Multi-view}
        & View Matching          & 216 & 4 & 33.3\% \\
        & Rotation Comparison    & 243 & 4 & 33.3\% \\
        & Post-Rotation View     & 192 & 5 & 25.0\% \\
        \bottomrule
    \end{tabular}
\end{table}

\section{Evaluation Protocol and Task Definitions}
\label{app:task_examples}

We specify the evaluation protocol and illustrate the eight OR-Bench tasks, grouped into five two-view tasks and three multi-view tasks. Rotation angles denote unsigned shortest azimuthal changes about the object's vertical axis; direction is provided explicitly in the text when required. Each display follows the image order, question format, and answer space used during evaluation. For closed-set tasks, target classes and correct answer positions are balanced, with option order varied where applicable to limit positional bias. All reported results are computed on the complete task sets summarized in Table~\ref{tab:benchmark_composition}.

\subsection{Scoring and Decoding}

\paragraph{Evaluation protocol.} We evaluate nine open-source and three proprietary VLMs zero-shot under fixed input/output formats. For closed-set tasks, responses are parsed against the option letters allowed by each instance and scored by exact agreement with the stored answer. Unparseable responses are recorded as \texttt{UNKNOWN} and counted as incorrect. For Exact Angle (Open), numeric values are extracted and degree formatting is normalized; the score requires exact agreement with the ground-truth angle. We report per-task accuracy, Macro-8 as the unweighted mean of the eight task accuracies, and Micro-8 as accuracy over all examples.

\paragraph{Decoding.} All benchmark evaluations use greedy decoding (\texttt{do\_sample=false}) with at most 32 new tokens. Temperature and top-$p$ are unset when sampling is disabled. Each model is evaluated once on the full benchmark; the reported results are single-run measurements.

\subsection{Two-view Tasks}

Five tasks query the rotation relation within a single image pair at different output granularities. 

\begin{taskbox}{Rotation Detection}

\begin{center}
\taskimage[0.35\linewidth]{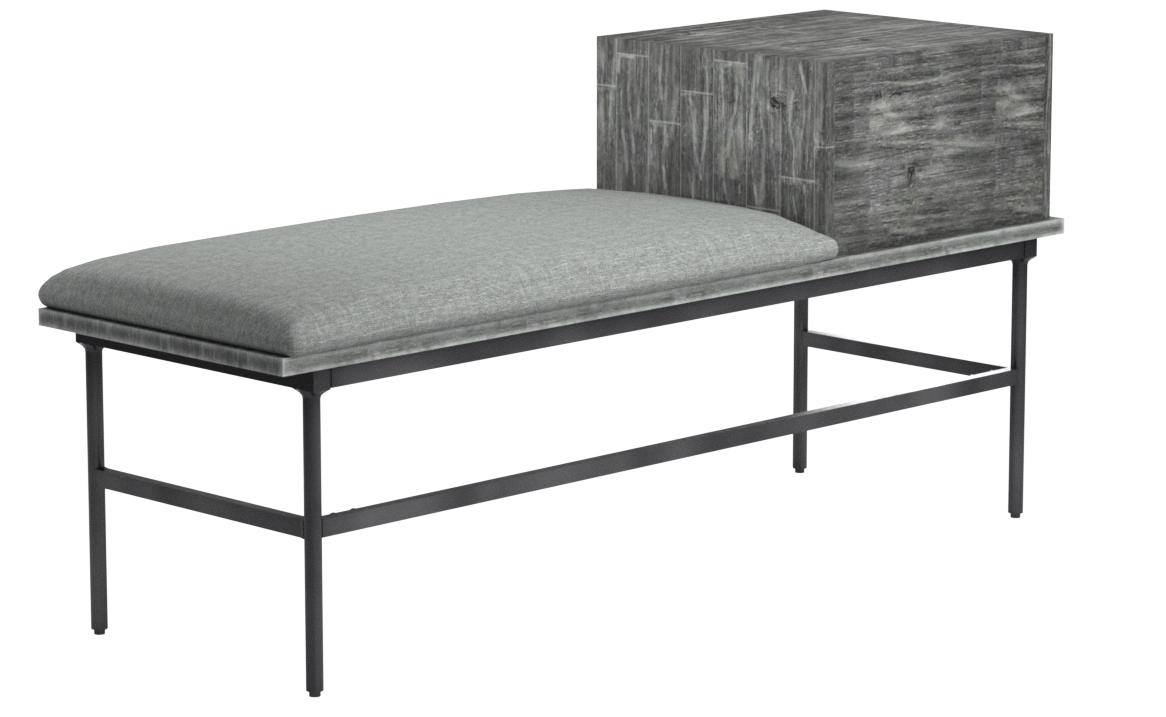}{Image 1}
\hspace{6mm}
\taskimage[0.35\linewidth]{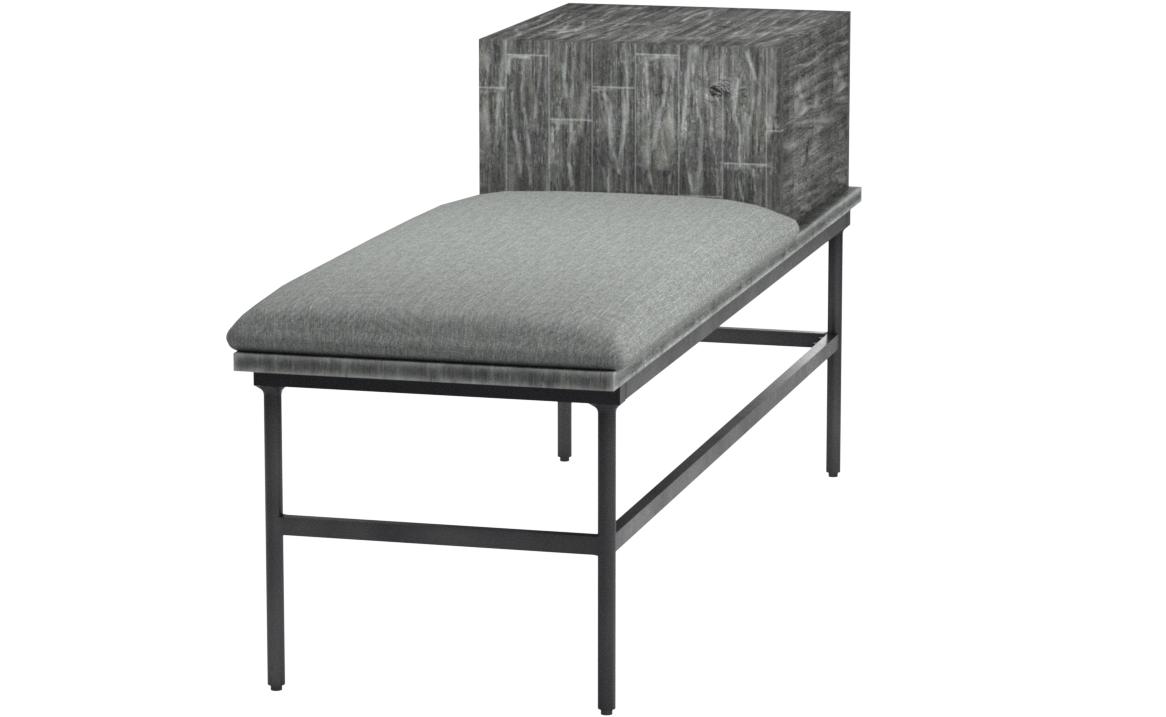}{Image 2}
\end{center}

\smallskip
\noindent
The camera is fixed and the object rotates in place.

\smallskip
\noindent
Do the two images show the object in the same orientation, or has it rotated?

\smallskip
\noindent
A. Rotated\\
B. Same orientation

\smallskip
\noindent
\textit{Only answer with a single capital letter from (A, B).}

\tcblower

\footnotesize
\textbf{Setup.} The label is \textit{Rotated} iff $\theta>0^\circ$; same-orientation examples reuse the identical frame. Target classes and correct answer positions are balanced.

\end{taskbox}

\begin{taskbox}{Coarse Angle}

\begin{center}
\taskimage[0.35\linewidth]{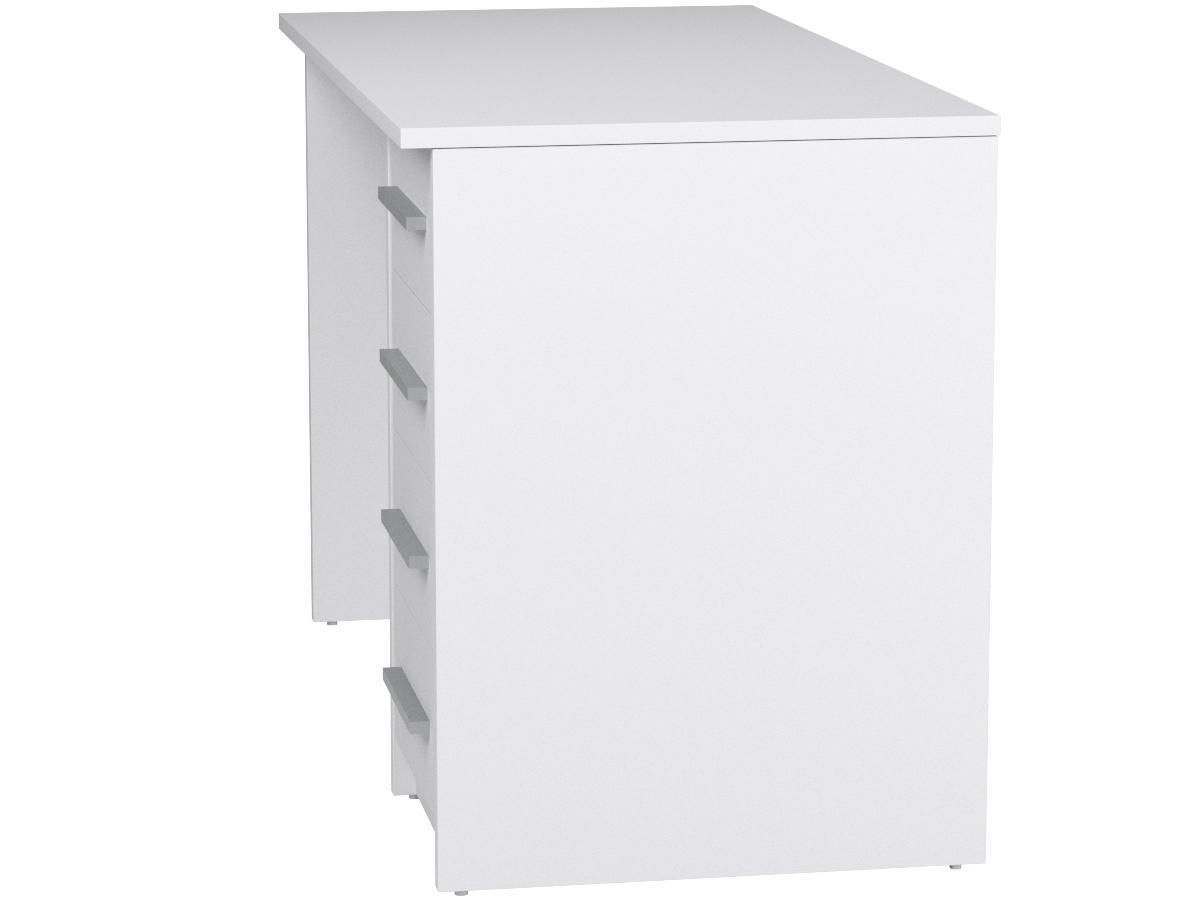}{Image 1}
\hspace{6mm}
\taskimage[0.35\linewidth]{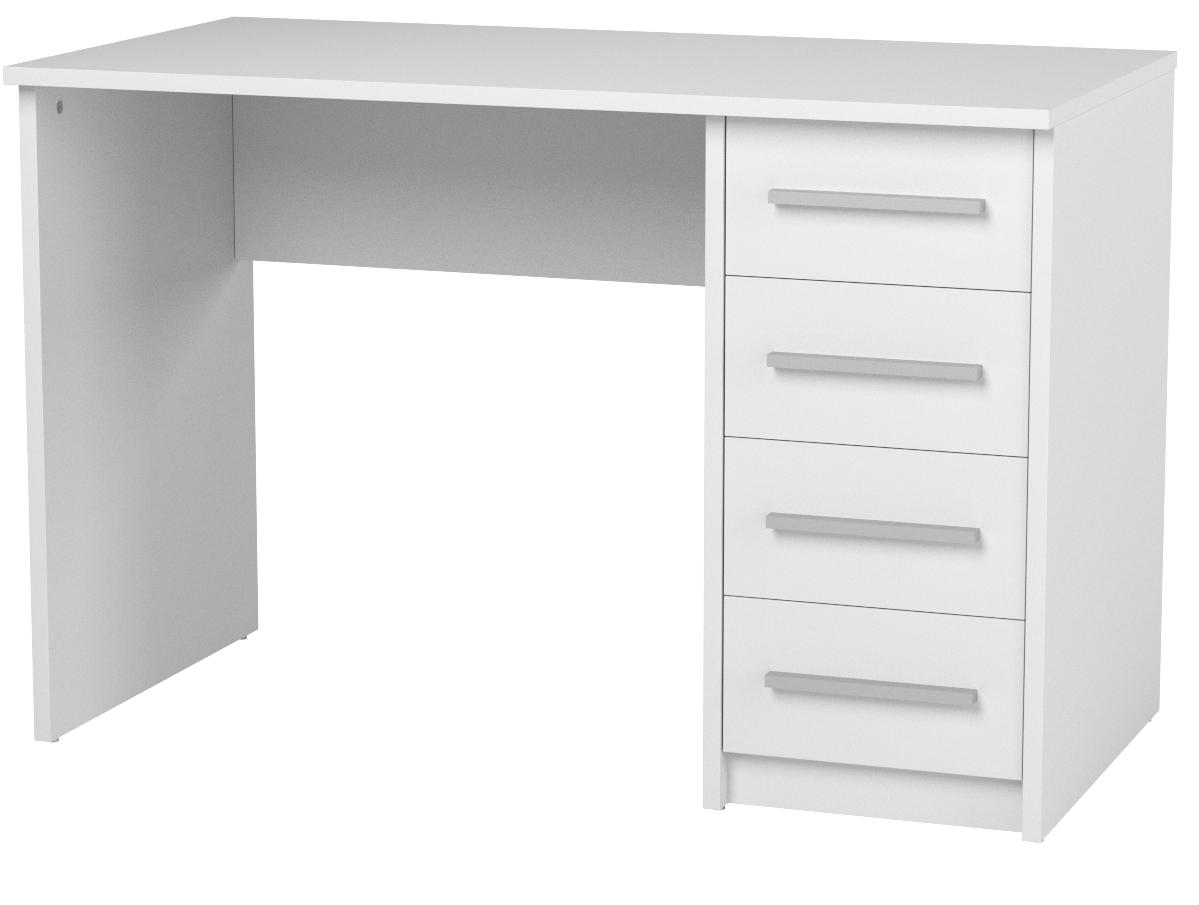}{Image 2}
\end{center}

\smallskip
\noindent
The camera is fixed and the object rotates in place.

\smallskip
\noindent
Which coarse interval best describes the rotation from Image 1 to Image 2?

\smallskip
\noindent
A. $[0^\circ,60^\circ)$\\
B. $[60^\circ,120^\circ]$\\
C. $(120^\circ,180^\circ]$

\smallskip
\noindent
\textit{Only answer with a single capital letter from (A, B, C).}

\tcblower

\footnotesize
\textbf{Setup.} The three intervals contain $\{0^\circ,30^\circ,45^\circ\}$, $\{60^\circ,90^\circ,120^\circ\}$, and $\{135^\circ,150^\circ,180^\circ\}$, respectively. Target classes and correct answer positions are balanced; option order varies across examples.

\end{taskbox}

\begin{taskbox}{$90^\circ$ Threshold}

\begin{center}
\taskimage[0.35\linewidth]{Figures/examples/0d52dd22_27.jpg}{Image 1}
\hspace{6mm}
\taskimage[0.35\linewidth]{Figures/examples/0d52dd22_33.jpg}{Image 2}
\end{center}

\smallskip
\noindent
The camera is fixed and the object rotates in place.

\smallskip
\noindent
Is the unsigned rotation angle from Image 1 to Image 2 less than, equal to, or greater than $90^\circ$?

\smallskip
\noindent
A. Greater than $90^\circ$\\
B. Equal to $90^\circ$\\
C. Less than $90^\circ$

\smallskip
\noindent
\textit{Only answer with a single capital letter from (A, B, C).}

\tcblower

\footnotesize
\textbf{Setup.} The three target classes are $\theta<90^\circ$, $\theta=90^\circ$, and $\theta>90^\circ$. Target classes and correct answer positions are balanced; option order varies across examples.

\end{taskbox}

\begin{taskbox}{Exact Angle (MCQ)}

\begin{center}
\taskimage[0.30\linewidth]{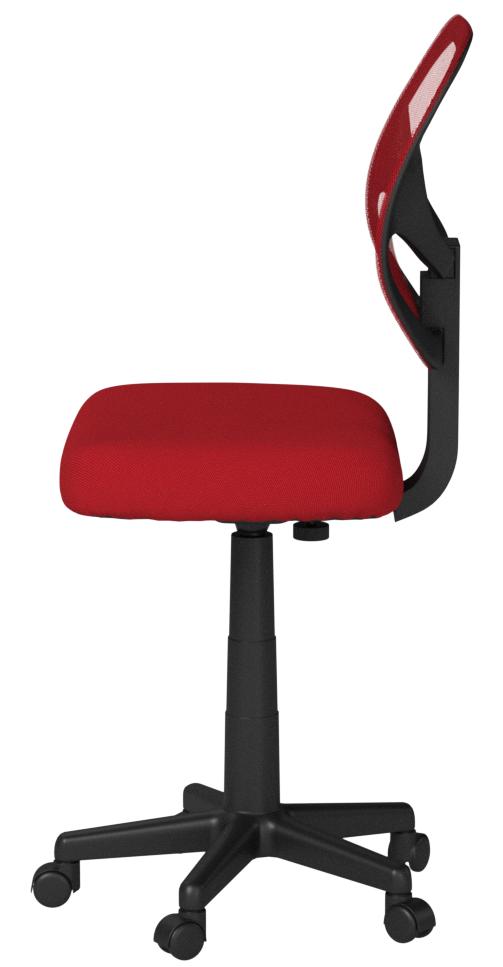}{Image 1}
\hspace{6mm}
\taskimage[0.30\linewidth]{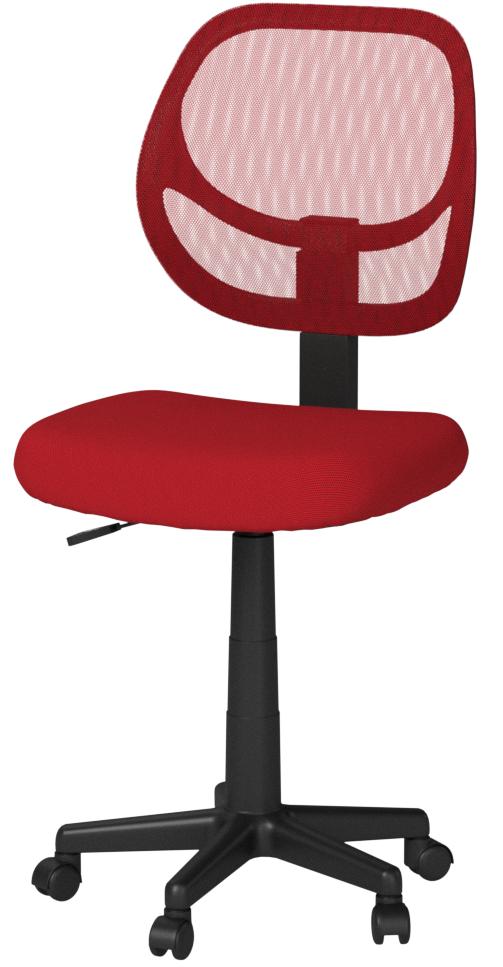}{Image 2}
\end{center}

\smallskip
\noindent
The camera is fixed and the object rotates in place.

\smallskip
\noindent
How many degrees did the object rotate from Image 1 to Image 2?

\smallskip
{\scriptsize
\noindent
\resizebox{\linewidth}{!}{%
\begin{tabular}{lllllllll}
A. $30^\circ$ & B. $45^\circ$ & C. $60^\circ$ & D. $90^\circ$ & E. $120^\circ$ & F. $135^\circ$ & G. $150^\circ$ & H. $180^\circ$ & I. $0^\circ$
\end{tabular}
}
}

\smallskip
\noindent
\textit{Only answer with a single capital letter from (A, B, C, D, E, F, G, H, I).}

\tcblower

\footnotesize
\textbf{Setup.} All nine reference angles occur equally often. Option order varies across examples, and correct answer positions are balanced.

\end{taskbox}

\begin{taskbox}{Exact Angle (Open)}

\begin{center}
\taskimage[0.30\linewidth]{Figures/examples/0a694a0e_24.jpg}{Image 1}
\hspace{6mm}
\taskimage[0.30\linewidth]{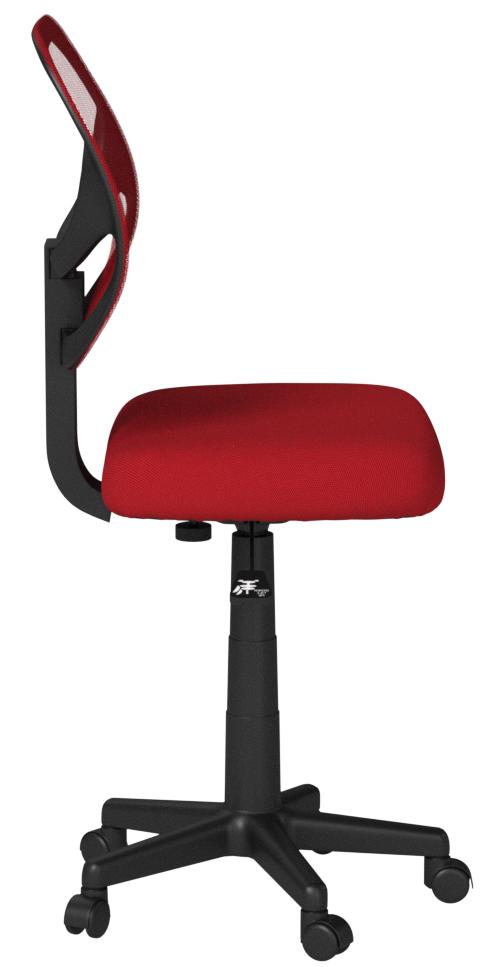}{Image 2}
\end{center}

\smallskip
\noindent
The camera is fixed and the object rotates in place.

\smallskip
\noindent
How many degrees did the object rotate from Image 1 to Image 2?

\smallskip
\noindent
\textit{Answer with a single number in degrees.}

\tcblower

\footnotesize
\textbf{Setup.} Exact Angle (Open) removes the candidate list and requires a direct numeric prediction. The score requires exact agreement with the ground-truth angle after normalization.

\end{taskbox}

\subsection{Multi-view Tasks}

Three tasks require models to match, compare, and apply rotation relations across multiple observations.

\begin{taskbox}{View Matching}

\begin{center}
\taskimage[0.22\linewidth]{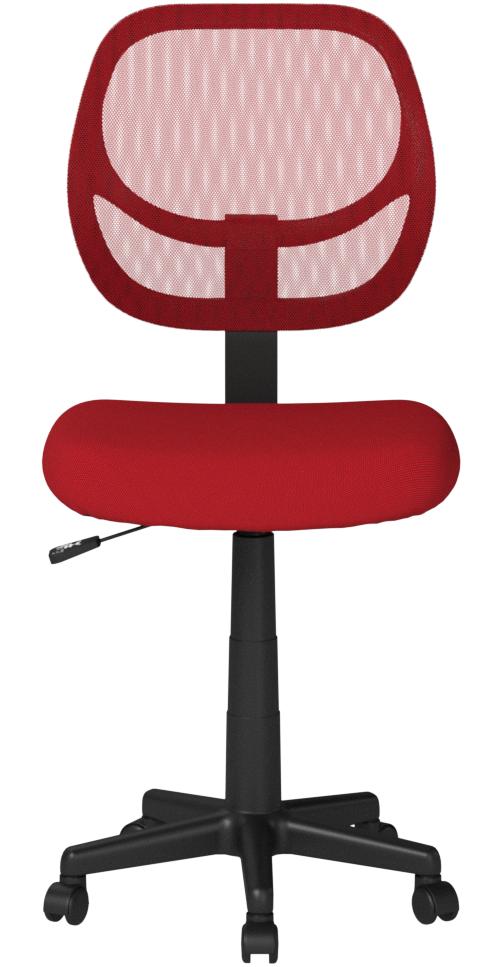}{Reference}
\hfill
\taskimage[0.22\linewidth]{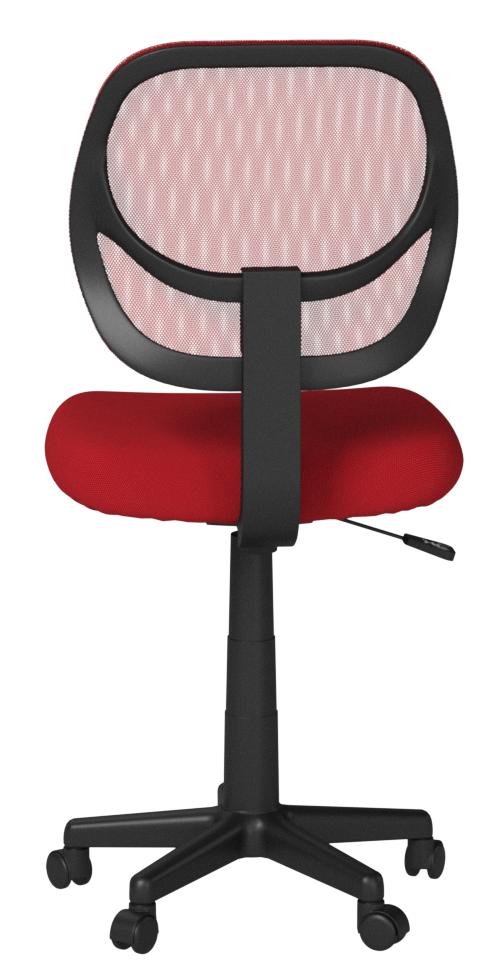}{Candidate \textbf{A}}
\hfill
\taskimage[0.22\linewidth]{Figures/examples/0a694a0e_24.jpg}{Candidate \textbf{B}}
\hfill
\taskimage[0.22\linewidth]{Figures/examples/0a694a0e_60.jpg}{Candidate \textbf{C}}
\end{center}

\smallskip
\noindent
The reference image shows the object from the \textbf{Front} view. Select the candidate image that best shows the object from the \textbf{Left} view.

\smallskip
\noindent
\textit{Only answer with a single capital letter from (A, B, C).}

\tcblower

\footnotesize
\textbf{Setup.} Reference and target cardinal views vary across examples. All candidates depict the same object as the reference, and correct answer positions are balanced.

\end{taskbox}

\begin{taskbox}{Rotation Comparison}

\smallskip
\noindent
The camera is fixed. Each pair shows the same object before and after an in-place rotation.

\begin{center}
\begin{minipage}[t]{0.46\linewidth}
    \centering
    \textbf{Pair A}\\[4pt]
    \taskimage[0.40\linewidth]{Figures/examples/0a694a0e_11.jpg}{Image 1}
    \hspace{4mm}
    \taskimage[0.40\linewidth]{Figures/examples/0a694a0e_06.jpg}{Image 2}
\end{minipage}
\hfill
\begin{minipage}[t]{0.46\linewidth}
    \centering
    \textbf{Pair B}\\[4pt]
    \taskimage[0.40\linewidth]{Figures/examples/0a694a0e_60.jpg}{Image 1}
    \hspace{4mm}
    \taskimage[0.40\linewidth]{Figures/examples/0a694a0e_11.jpg}{Image 2}
\end{minipage}
\end{center}

\smallskip
\noindent
Which pair has the larger unsigned rotation angle?

\smallskip
\noindent
A. The two pairs have the same rotation angle.\\
B. Pair A has the larger rotation.\\
C. Pair B has the larger rotation.

\smallskip
\noindent
\textit{Only answer with a single capital letter from (A, B, C).}

\tcblower

\footnotesize
\textbf{Setup.} The target is determined directly by comparing $\theta_A$ and $\theta_B$. All four images depict the same physical object, preventing category or object identity from serving as a shortcut. The three comparison outcomes and correct answer positions are balanced.

\end{taskbox}

\begin{taskbox}{Post-Rotation View}

\smallskip
\noindent
Study the reference image, then mentally simulate the object rotating in place.

\begin{center}
\taskimage[0.30\linewidth]{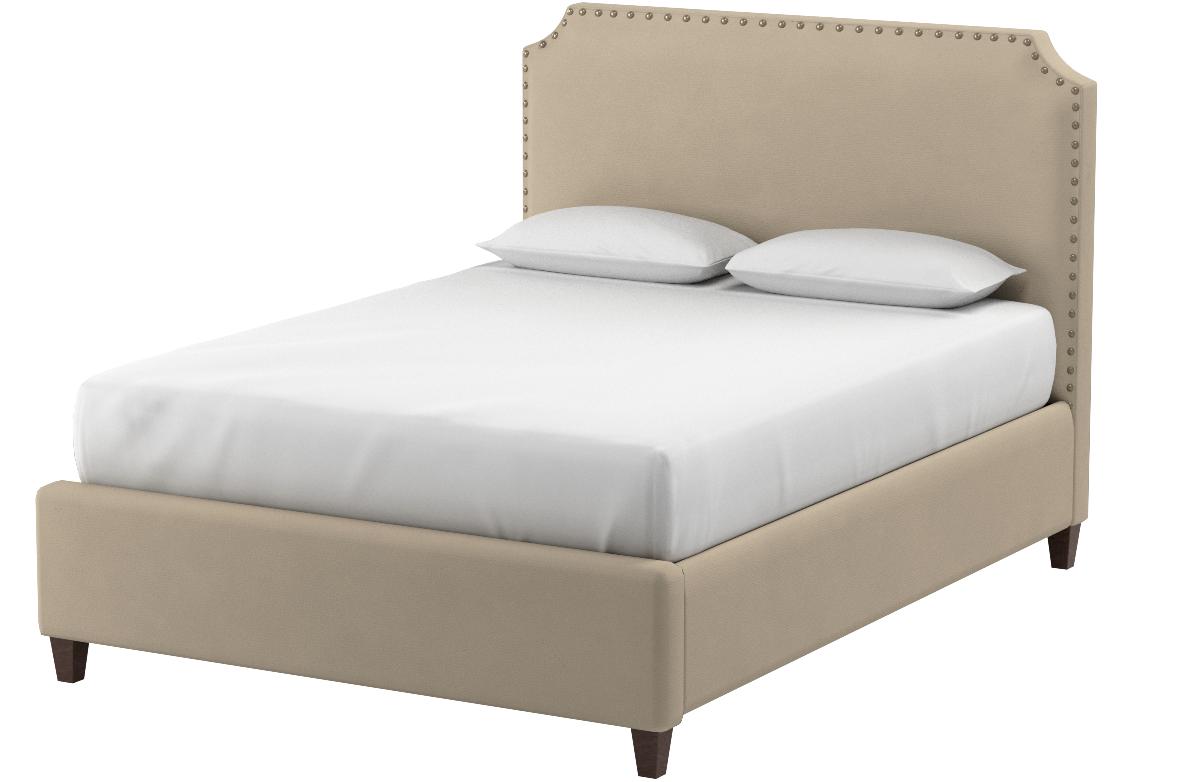}{Reference}
\end{center}

\smallskip
\noindent
Imagine the object rotates \textbf{$150^\circ$ clockwise} while the camera stays fixed. Clockwise and counterclockwise are defined from a top-down view.

\smallskip
\noindent
After imagining this rotation, which candidate image matches the final view?

\smallskip
\begin{center}
\taskimage[0.21\linewidth]{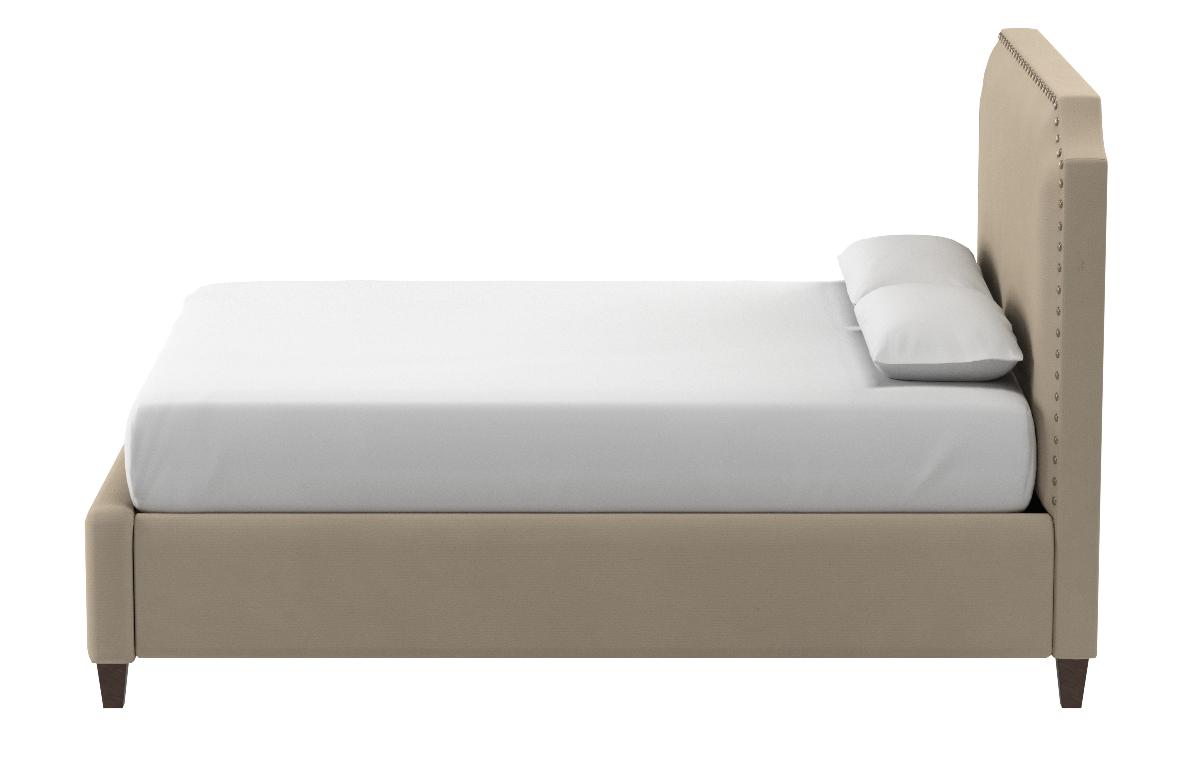}{Candidate A}
\hfill
\taskimage[0.21\linewidth]{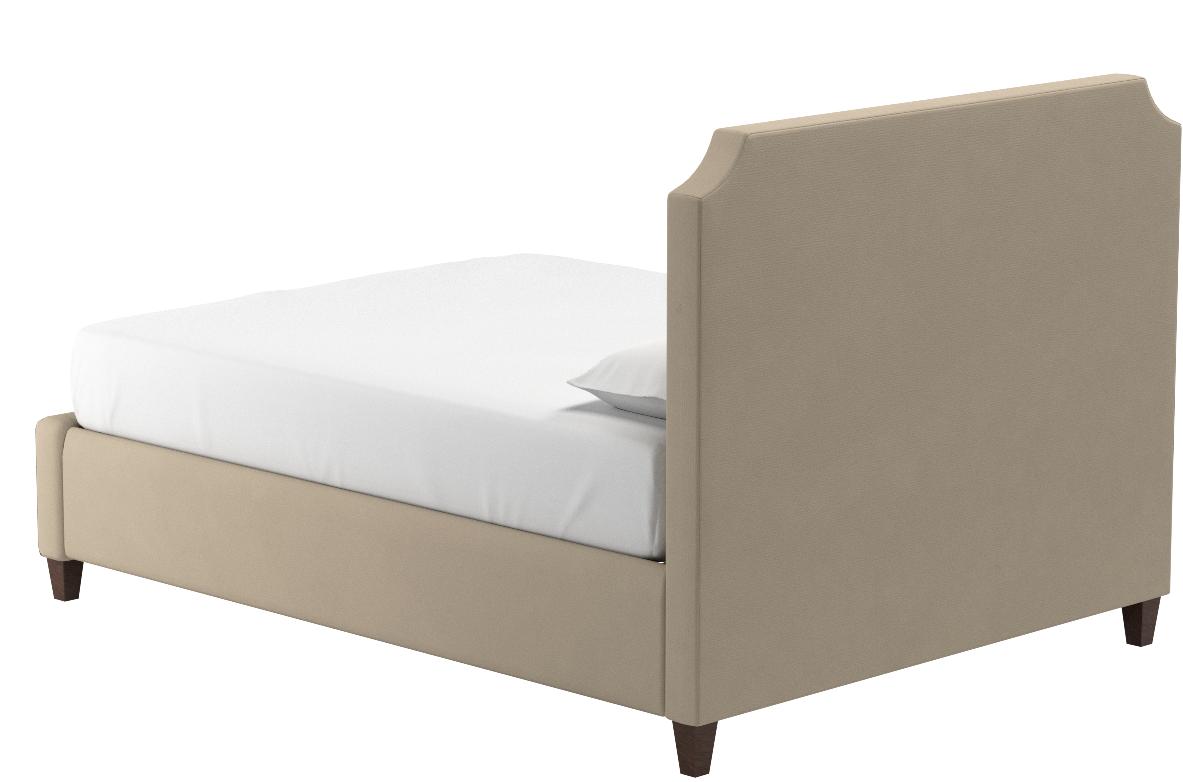}{Candidate B}
\hfill
\taskimage[0.21\linewidth]{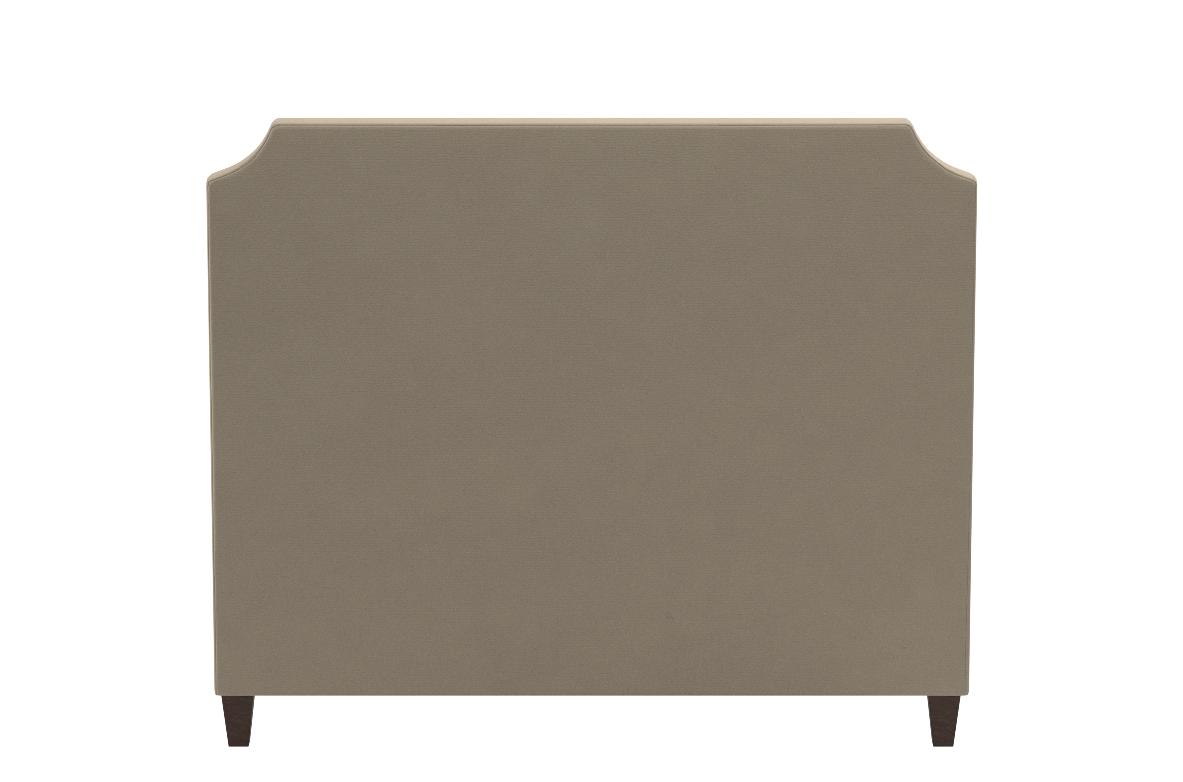}{Candidate C}
\hfill
\taskimage[0.21\linewidth]{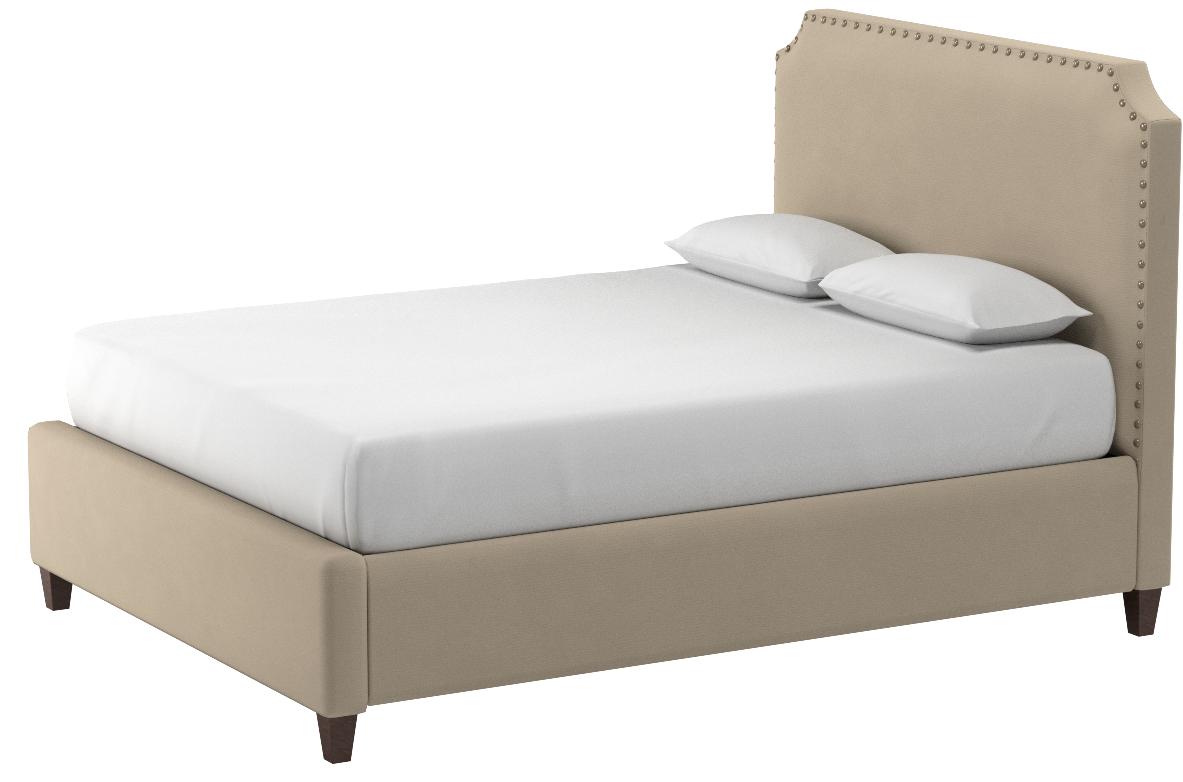}{Candidate D}
\end{center}

\smallskip
\noindent
\textit{Only answer with a single capital letter from (A, B, C, D).}

\tcblower

\footnotesize
\textbf{Setup.} Rotation direction is explicitly specified in the instruction. All candidates depict the same object and are sampled from the instructed-direction trajectory. Duplicate candidate images and opposite-direction, same-magnitude distractors are excluded, and correct answer positions are balanced.

\end{taskbox}

\subsection{Output-Format Controls}

We perform two checks for output-format artifacts. First, all 486 Rotation Detection responses from each of Qwen2-VL-7B, LLaVA-OneVision-Qwen2-7B, and Cosmos-Reason1-7B are successfully parsed, despite accuracies of 62.8\%, 60.5\%, and 65.0\%, respectively. Second, the option preferred by the LM head agrees with the option recovered from the final generated response in 100\% of evaluated cases. In these checks, errors are not explained by response parsing or a mismatch between native option readout and the emitted answer.

\section{Human Reference Evaluation}
\label{app:human_eval}

Human performance is measured on a stratified subset of OR-Bench as a reference for interpreting model performance. Five participants each complete 628 sampled examples, yielding 3,140 participant--item responses in total. The subset covers all eight tasks: 96 examples for Rotation Detection, 81 for Coarse Angle, 72 for the $90^\circ$ Threshold, 81 each for Exact Angle (MCQ) and Exact Angle (Open), 72 for View Matching, 81 for Rotation Comparison, and 64 for Post-Rotation View. Sampling is stratified by task-relevant variables, including rotation angle, target class, comparison outcome, and correct answer position, to provide balanced coverage of each task.

\begin{figure}[!ht]
    \centering
    \includegraphics[width=\linewidth]{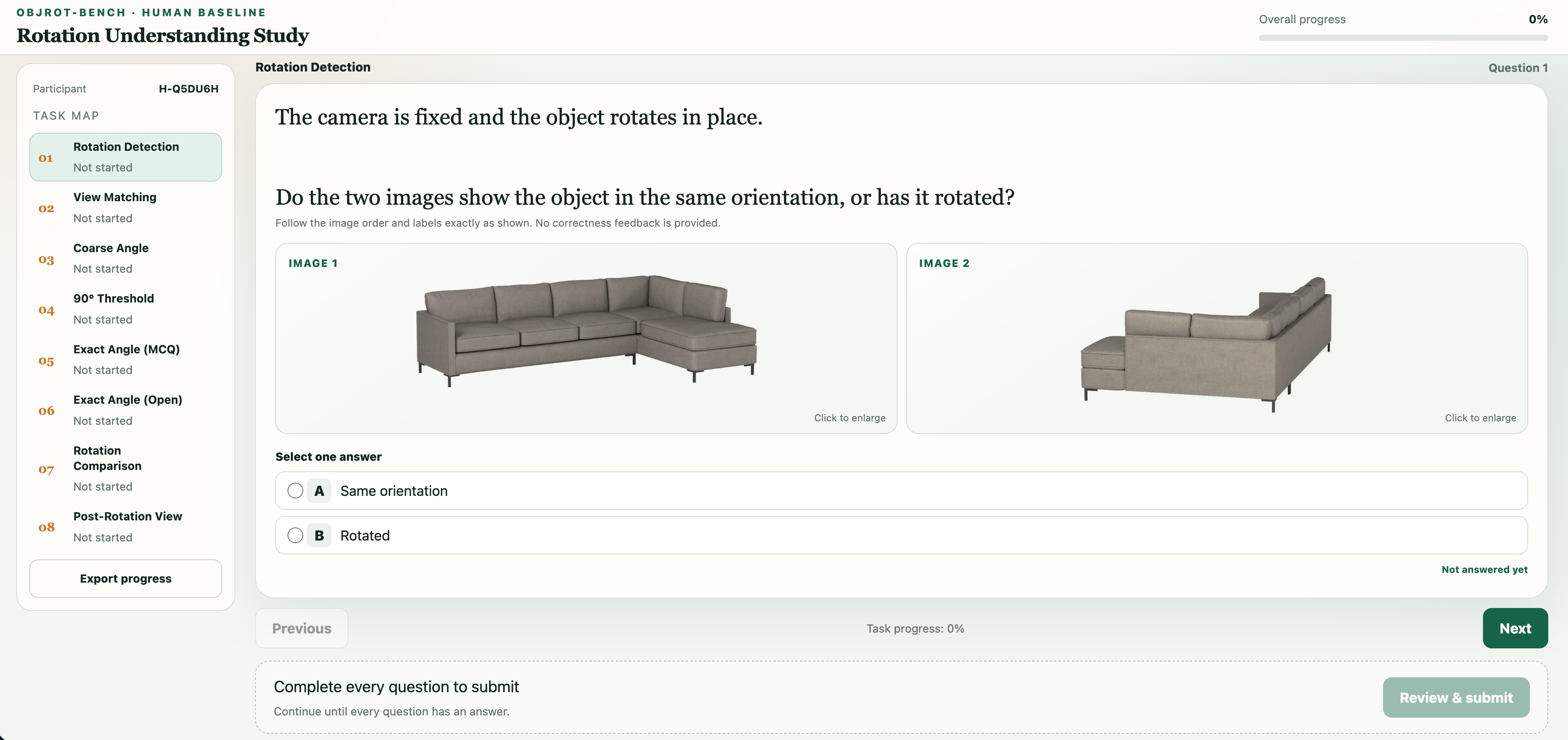}
    \caption{\textbf{Human evaluation interface for OR-Bench.} The interface supports task-wise navigation and progress tracking. A Rotation Detection example is shown.}
    \label{fig:anno}
\end{figure}

Evaluation is divided into four sessions without correctness feedback. Participants receive the same images, task instructions, questions, and answer spaces used for model evaluation. Human responses are scored using the same task-level criteria as model outputs, including normalized exact match for Exact Angle (Open). Human Macro-8 is the unweighted mean of accuracy across the eight tasks.

\section{Canonical-Angle Collapse in Proprietary Models}
\label{sec:appendix_proprietary_collapse}

Proprietary models also concentrate their predictions on the canonical angles $\{0^\circ,90^\circ,180^\circ\}$, consistent with the pattern described in the main text (Table~\ref{tab:proprietary_collapse}). The most frequently predicted angle is $90^\circ$ for all three models.

\begin{table}[!ht]
    \centering
    \caption{Share of predictions assigned to canonical angles and to $90^\circ$ by proprietary models (\%).}
    \label{tab:proprietary_collapse}
    \small
    \setlength{\tabcolsep}{12pt}
    \renewcommand{\arraystretch}{1.12}
    \begin{tabular}{@{}lrr@{}}
        \toprule
        Model & $\{0^\circ,90^\circ,180^\circ\}$ & $90^\circ$ \\
        \midrule
        GPT-4o & 82.0 & 62.3 \\
        GPT-4o-mini & 98.4 & 70.5 \\
        GPT-5.4 & 68.9 & 28.7 \\
        \bottomrule
    \end{tabular}
\end{table}

\section{Representation Diagnostics: Probe Design and Implementation}
\label{sec:appendix_probe}

We use linear probes to measure task-relevant rotation information recoverable from frozen VLM representations. The main analysis covers three increasingly fine-grained tasks: Rotation Detection, Coarse Angle, and Exact Angle. We specify the task alignment, object-held-out protocol, evaluation metric, and two readout locations, followed by the additional $90^\circ$ Threshold results and random-label control.

\subsection{Task Alignment and Object-Held-Out Protocol}
\label{sec:appendix_probe_protocol}

We probe the representations using the examples and labels of the corresponding OR-Bench tasks: Rotation Detection (486 examples), Coarse Angle (243), and Exact Angle (243). These tasks progressively increase the angular resolution required by the readout.

All probes use five-fold object-disjoint cross-validation. Let \(\mathcal{O}_{\mathrm{train}}^{(k)}\) and \(\mathcal{O}_{\mathrm{test}}^{(k)}\) denote the training and test objects in fold \(k\). We enforce
\[
\mathcal{O}_{\mathrm{train}}^{(k)} \cap \mathcal{O}_{\mathrm{test}}^{(k)} = \varnothing.
\]
Each example receives exactly one out-of-fold prediction, and all preprocessing and classifier fitting use only the corresponding training fold.

\subsection{Evaluation Metric}
\label{sec:appendix_probe_metric}

We compute balanced accuracy (BA):
\[
\operatorname{BA} = \frac{1}{C}\sum_{c=1}^{C}\operatorname{Recall}_c,
\]
where \(C\) is the number of classes. BA gives equal weight to each class.

To account for the different uniform-reference levels across tasks, we report reference-adjusted balanced accuracy:
\[
\operatorname{AdjBA} = 100 \times \frac{\operatorname{BA}-q}{1-q}, \qquad q=\frac{1}{C}.
\]
Under this convention, \(\operatorname{AdjBA}=0\) corresponds to uniform guessing and \(\operatorname{AdjBA}=100\) to perfect classification; negative values indicate performance below the uniform reference. This places the 2-way, 3-way, and 9-way probes on a common reference-adjusted scale.

\subsection{Vision-Encoder Readout}
\label{sec:appendix_ve_probe}

We examine three depths of the frozen vision encoder: L8, L16, and the final layer. For each image pair, tokens associated with Image 1 and Image 2 are separated and independently mean-pooled:
\[
f_1=\operatorname{MeanPool}(F_1), \qquad f_2=\operatorname{MeanPool}(F_2).
\]
Mean pooling introduces no learned aggregation parameters.

For the primary vision-encoder (VE) probe, the pair is represented by the parameter-free absolute difference:
\[
%r_{\mathrm{VE}} = \operatorname{absdiff}(f_1,f_2) = \lvert f_1-f_2\rvert.
\lvert f_1-f_2\rvert.
\]
This representation is symmetric to image order and aligned with the unsigned rotation quantities used by the probe tasks. It introduces no learnable relational module, so the recovered rotation information is carried by the frozen single-image representations.

The readout uses feature standardization followed by a linear RidgeClassifier:
\[
\lvert f_1-f_2\rvert \longrightarrow \operatorname{StandardScaler} \longrightarrow \operatorname{RidgeClassifier}(\alpha=10).
\]
The scaler and classifier are fitted independently within each training fold, while all VLM parameters remain frozen.

We additionally evaluate signed difference, concatenation, and richer pair representations as diagnostic variants. These serve as robustness checks; the primary VE analysis uses absolute difference throughout.

\subsection{Matched Readout at the Native Answer Interface}
\label{sec:appendix_hpre_probe}

To test whether the VLM itself has constructed a two-view relation, we probe the state immediately consumed by its native LM head.

We denote by \(h_{\mathrm{pre}}\) the final post-normalization hidden state used to predict the first answer token. At this location, no external absolute difference, pooling operation, or pair representation is introduced. The probe therefore tests whether the two-view rotation relation is linearly readable from the model's own representation.

Both readouts operate on the same hidden state. Native prediction uses the frozen vocabulary head:
\[
\ell_{\mathrm{native}} = W_{\mathrm{vocab}}h_{\mathrm{pre}},
\]
whereas the task-aligned probe uses fitted linear weights:
\[
\ell_{\mathrm{probe}} = W_{\mathrm{probe}}h_{\mathrm{pre}}.
\]
The comparison tests whether task-relevant rotation information reaches the native answer interface and whether the pretrained readout effectively uses it.

\subsection{Additional Probing Results: Task and Protocol Generalization}
\label{sec:appendix_threshold_probe}

We report two robustness checks on the coarse-granularity readout gap: one
varying the task, and one varying the probe training protocol.

\paragraph{A different task and question form.}
The $90^\circ$ Threshold task uses a separate image set and a different question
form from Coarse Angle. All three models answer at chance on it
(5.6--6.2 AdjBA). Linear probes nonetheless recover the rotation relation well
above chance at every VE layer, and at \(h_{\mathrm{pre}}\) the probe exceeds the
model's own answer for all three models (Table~\ref{tab:threshold_probe}). The
readout gap therefore reappears under a different image set and question form,
here on a task that no model solves.

\begin{table}[!ht]
    \centering
    \caption{Probe and native performance on $90^\circ$ Threshold
    (AdjBA, \%; chance-adjusted, chance $=0$). VE probes use \texttt{absdiff} on
    the two frozen single-image representations; the \(h_{\mathrm{pre}}\) probe
    and the LM head read the same representation. \emph{Native} is the model's
    own answer and \(\Delta\) is the probe--native difference at
    \(h_{\mathrm{pre}}\). Evaluated on 243 image pairs.}
    \label{tab:threshold_probe}
    \small
    \setlength{\tabcolsep}{7pt}
    \renewcommand{\arraystretch}{1.12}
    \begin{tabular}{@{}lrrrrrr@{}}
        \toprule
        & \multicolumn{3}{c}{VE probes} & \multicolumn{3}{c}{Output-head readout} \\
        \cmidrule(lr){2-4}
        \cmidrule(l){5-7}
        Model & L8 & L16 & Final & \(h_{\mathrm{pre}}\) & LM Head & \(\Delta\) \\
        \midrule
        Qwen3-VL-8B    & 38.4 & 22.2 & 13.4 & 27.2 & 5.6 & $+21.6$ \\
        Qwen2.5-VL-7B  & 29.2 & 31.0 & 13.9 &  9.9 & 6.2 & $+3.7$  \\
        InternVL3.5-8B & 31.0 & 20.8 & 24.1 & 24.1 & 6.2 & $+17.9$ \\
        \bottomrule
    \end{tabular}
\end{table}

\paragraph{A different probe training protocol.}
To verify that the probes do not rely on idiosyncrasies of the evaluation set,
we refit them entirely on external data. For each frozen model and each task we
train an independent three-way linear probe on 972 ALOI image pairs, with 243
pairs held out for validation, object-disjoint splits, and angle, class and
option distributions matched to the target task. We use the native task prompt,
extract \(h_{\mathrm{pre}}\) at the last prompt position, standardize with
training-set statistics, and fit a \texttt{RidgeClassifier} whose penalty is
selected on the external validation set only. The OR-Bench test sets (243 pairs
each) are used for evaluation and never for fitting or model selection.

Under this protocol the probe again exceeds the model's own answer on both tasks
and all three models (Table~\ref{tab:external_probe}). Since the probe is
trained on a different image distribution, these values are a lower bound on the
linearly available rotation information.

\begin{table}[!ht]
    \centering
    \caption{Probes trained on external ALOI data and evaluated on OR-Bench
    (AdjBA, \%; chance-adjusted, chance $=0$). Probes are fit on 972 ALOI pairs
    with object-disjoint splits and never see the test set. \(\Delta\) is the
    probe--native difference. Each test set contains 243 image pairs.}
    \label{tab:external_probe}
    \small
    \setlength{\tabcolsep}{7pt}
    \renewcommand{\arraystretch}{1.12}
    \begin{tabular}{@{}lrrrrrr@{}}
        \toprule
        & \multicolumn{3}{c}{Coarse Angle} & \multicolumn{3}{c}{$90^\circ$ Threshold} \\
        \cmidrule(lr){2-4}
        \cmidrule(l){5-7}
        Model & Probe & Native & \(\Delta\) & Probe & Native & \(\Delta\) \\
        \midrule
        Qwen3-VL-8B    & 22.2 & 11.1   & $+11.1$ & 32.1 & 5.6 & $+26.5$ \\
        Qwen2.5-VL-7B  & 16.0 & 12.3   & $+3.7$  & 21.0 & 6.2 & $+14.8$ \\
        InternVL3.5-8B & 21.0 & -1.9 & $+22.8$ & 14.2 & 6.2 & $+8.0$  \\
        \bottomrule
    \end{tabular}
\end{table}

Across both checks, the gap holds in every model and task examined: more
rotation information is linearly present in the representation than the model's
own answer reports.
\subsection{Random-Label Control}
\label{sec:appendix_random_control}

We perform a random-label control on Coarse Angle using the VE L8 representations. For each fold, we keep the object split, frozen features, preprocessing, and Ridge configuration unchanged, but randomly permute the labels within the training partition:
\[
\tilde{y}_{\mathrm{train}} = \pi(y_{\mathrm{train}}),
\]
where \(\pi\) is a random permutation. Held-out labels remain unchanged. Randomizing the training labels reduces AdjBA below the uniform reference for all three models, supporting that the recovered signal reflects rotation information associated with the true labels.

\begin{table}[!ht]
    \centering
    \caption{Random-label control for Coarse Angle readout from VE L8 representations (AdjBA, \%).}
    \label{tab:random_label_control}
    \small
    \setlength{\tabcolsep}{12pt}
    \renewcommand{\arraystretch}{1.15}
    \begin{tabular}{@{}lcc@{}}
        \toprule
        \textbf{Model} & \textbf{True labels} & \textbf{Random labels} \\
        \midrule
        Qwen3-VL-8B & 35.8 & \(-6.2\) \\
        Qwen2.5-VL-7B & 28.4 & \(-4.3\) \\
        InternVL3.5-8B & 38.9 & \(-3.8\) \\
        \bottomrule
    \end{tabular}
\end{table}

\section{RotationCue: Recovery, Training, and Intervention Details}
\label{sec:appendix_rotationcue}

RotationCue recovers a structured rotation state from frozen internal visual representations and reinjects that state as context for the original VLM. No parameter of the base VLM is updated at any stage. For two images \(I_A\) and \(I_B\), we independently extract mean-pooled Vision L8 representations,
\[
F_A =
\operatorname{MeanPool}
\left(
V^{(8)}(I_A)
\right),
\qquad
F_B =
\operatorname{MeanPool}
\left(
V^{(8)}(I_B)
\right).
\]

An external decoder \(D_\phi\) predicts the rotation state,
\[
\hat z = D_\phi\bigl(R_{\mathrm{full}}(F_A,F_B)\bigr),
\]
which is serialized as fixed structured context \(C(\hat z)\) and supplied to the unchanged base model,
\[
\hat y
=
M_\theta
\bigl(
I_A,I_B,q,C(\hat z)
\bigr),
\qquad
\theta \ \text{fixed}.
\]

Thus, RotationCue uses the VLM only as a source of frozen internal representations and as the final reasoner; only the decoder parameters \(\phi\) are trained.

The section proceeds from the predicted state and frozen visual inputs to the decoder architecture and hierarchical objective. It then documents the independent training data, shared optimization protocol, multi-image pairing, state-injection controls, and robustness evaluation.

\subsection{Structured Rotation State}
\label{sec:appendix_rotationcue_state}

RotationCue predicts a hierarchical state
\[
z=(z_s,z_r,z_c),
\]
containing three fields. The first,
\[
z_s\in\{\texttt{yes},\texttt{no}\},
\]
indicates whether the two images depict the same object. Conditional on \(z_s=\texttt{yes}\),
\[
z_r\in\{\texttt{yes},\texttt{no}\}
\]
indicates whether its orientation changes. When rotation is present,
\[
z_c\in
\{\texttt{small},\texttt{medium},\texttt{large}\}
\]
describes its coarse magnitude,
\[
\texttt{small}:(0^\circ,60^\circ),
\qquad
\texttt{medium}:[60^\circ,120^\circ],
\qquad
\texttt{large}:(120^\circ,180^\circ].
\]

The state follows the dependency
\[
z_s \rightarrow z_r \rightarrow z_c.
\]

If the images depict different objects, the rotation relation is undefined. When no rotation occurs, \texttt{angle\_range} is set to \texttt{not\_applicable}. For same-object pairs, \texttt{rotation\_present=no} corresponds to $\theta=0^\circ$. Together, this no-rotation state and the positive \texttt{small} angle range cover the benchmark's \texttt{small} class $[0^\circ,60^\circ)$.

We recover a coarse intermediate state because Coarse Angle retains substantial recoverable information while showing the largest gap from the model's own predictions. RotationCue supplies this state as context for the original VLM to combine with the images and task requirements.

\subsection{Frozen Visual Representation and Pair Construction}
\label{sec:appendix_rotationcue_pair}

The two images are encoded independently. RotationCue uses only frozen Vision L8 representations and does not use language-model hidden states.

From \(F_A\) and \(F_B\), we construct the fixed pairwise relation representation
\[
R_{\mathrm{full}}
=
\left[
F_A,\,
F_B,\,
F_B-F_A,\,
\lvert F_B-F_A\rvert,\,
F_A\odot F_B
\right].
\]

The five components preserve the individual views, their signed change, unsigned discrepancy, and element-wise agreement.

The diagnostic probe uses the \(\operatorname{absdiff}\) representation, whereas RotationCue uses all five components to recover the structured state. Pair construction is fixed and introduces no learned parameters before the decoder.

\subsection{Decoder Architecture}
\label{sec:appendix_rotationcue_arch}

Let \(d_m\) denote the Vision L8 feature dimension of base model \(m\). Each component of \(R_{\mathrm{full}}\) is projected into a shared 512-dimensional latent space,
\[
h_j=P_j(r_j),
\qquad
h_j\in\mathbb{R}^{512},
\]
using a Linear--GELU--Dropout projection with normalization as appropriate. Learned component-type embeddings distinguish the five relation components.

The decoder contains three learnable pose queries,
\[
Q^{(0)}
=
[q_s,q_r,q_c],
\]
corresponding to same-object, rotation-presence, and angle-range prediction. Two pre-layer-normalized extractor blocks update the queries,
\[
Q^{(\ell+1)}
=
\operatorname{PoseBlock}
\bigl(Q^{(\ell)},H\bigr),
\qquad
\ell\in\{0,1\}.
\]

Each block applies cross-attention from the state queries to the relation tokens, self-attention among the queries, and a \(512\rightarrow2048\rightarrow512\) feed-forward network. Attention uses eight heads with dropout \(0.1\).

The resulting query states are mapped to the three predictions,
\[
s=g_s(q_s),
\qquad
r=g_r(q_r),
\qquad
c=g_c(q_c),
\]
where each classifier uses
\[
\operatorname{LayerNorm}
\rightarrow
\operatorname{Linear}(512,512)
\rightarrow
\operatorname{GELU}
\rightarrow
\operatorname{Dropout}
\rightarrow
\operatorname{Linear}.
\]

Separate state queries preserve the semantic decomposition of the hierarchical relation rather than reducing it to a single flat class.

\subsection{Hierarchical Supervision}
\label{sec:appendix_rotationcue_loss}

Let \(y_s\), \(y_r\), and \(y_c\) denote the ground-truth same-object, rotation-presence, and angle-range labels. We define
\[
m_r
=
\mathbb{1}
[y_s=\texttt{yes}],
\qquad
m_c
=
\mathbb{1}
[y_s=\texttt{yes}\ \land\ y_r=\texttt{yes}],
\]
and optimize
\[
\mathcal{L}_{\mathrm{base}}
=
\operatorname{CE}(s,y_s)
+
\operatorname{CE}_{m_r}(r,y_r)
+
\operatorname{CE}_{m_c}(c,y_c).
\]

The masks prevent supervision of undefined states:
\[
y_s=\texttt{no}
\Rightarrow
y_r,y_c \ \text{undefined},
\]
and
\[
y_r=\texttt{no}
\Rightarrow
y_c \ \text{undefined}.
\]

We additionally use an expected angular-cost term,
\[
\mathcal{L}
=
\mathcal{L}_{\mathrm{base}}
+
\lambda_{\mathrm{angle}}
\mathcal{L}_{\mathrm{angle}},
\]
where
\[
\mathcal{L}_{\mathrm{angle}}
=
\frac{1}{\sum_i m_{c,i}}
\sum_i
m_{c,i}
\sum_{b=1}^{3}
p_{i,b}
\min
\left(
2,
\frac{\lvert\theta_i-a_b\rvert}{45}
\right),
\]
\(p_{i,b}\) is the predicted probability of angle range \(b\), and \(a_b\) is its predefined angular reference; angles in this expression are measured in degrees. We use
\[
\lambda_{\mathrm{angle}}=0.3.
\]

Ordinary cross-entropy treats all incorrect ranges equally. The angular term instead assigns a distance-dependent, capped cost to the predicted ranges, preserving the ordered geometry of the three ranges.

\subsection{Training Data and Object-Disjoint Splits}
\label{sec:appendix_rotationcue_data}

The decoder is trained exclusively on independent multi-view data: ALOI-View~\citep{geusebroek2005aloi} and SelectiveMV ModelNet40v1~\citep{alzahrani2024selectivemv,wu2015shapenets}.

ALOI contains 1,000 objects observed at 72 azimuths,
\[
0^\circ,5^\circ,\ldots,355^\circ.
\]

For two views with azimuths \(a\) and \(b\), the unsigned rotation magnitude is
\[
\theta
=
\min
\left(
|a-b|,
360^\circ-|a-b|
\right).
\]

SelectiveMV contains 3,983 objects with 12 RGB views. Each view is assigned
\[
a
=
30^\circ\times\texttt{view\_id}.
\]

Splits are constructed at the object level rather than the pair level,
\[
\mathcal{O}_{\mathrm{train}}
\cap
\mathcal{O}_{\mathrm{val}}
=
\varnothing,
\]
preventing different view pairs of the same physical object from entering both partitions.

OR-Bench is used only for final evaluation. Its objects, images, labels, questions, answer choices, and answer formats are not used for decoder training or checkpoint selection.

\subsection{Frozen Feature Caching and Shared Training Protocol}
\label{sec:appendix_rotationcue_training}

Vision L8 representations are extracted once and cached. The decoder is trained on the cached feature pairs while the base VLM remains frozen.

For all three VLMs, the decoder input is the mean-pooled Vision L8 representation. Feature extraction follows the native image interface of each vision backbone, while the decoder design and optimization protocol are kept fixed.

\begin{table}[!ht]
\centering
\caption{Frozen internal representations used as RotationCue inputs.}
\label{tab:rotationcue_features}
\small
\setlength{\tabcolsep}{6pt}
\renewcommand{\arraystretch}{1.12}
\begin{tabular}{@{}lll@{}}
\toprule
\textbf{Base VLM}
& \textbf{Decoder input}
& \textbf{Image preprocessing} \\
\midrule
Qwen3-VL-8B
& Vision L8 mean
& Qwen AutoProcessor, dynamic resolution \\
Qwen2.5-VL-7B
& Vision L8 mean
& Qwen AutoProcessor, dynamic resolution \\
InternVL3.5-8B
& Vision L8 mean
& \(448\times448\), at most one tile \\
\bottomrule
\end{tabular}
\end{table}

The two images are encoded independently in all cases, and no language-model hidden state is used as a decoder input.

Across all three VLMs, we use the same readout layer, relation representation, decoder design, and training protocol. A single decoder per base model serves all eight OR-Bench tasks. The shared configuration is summarized below.

\begin{table}[!ht]
\centering
\caption{Shared RotationCue decoder training configuration.}
\label{tab:rotationcue_training}
\small
\setlength{\tabcolsep}{10pt}
\renewcommand{\arraystretch}{1.12}
\begin{tabular}{@{}lc@{}}
\toprule
\textbf{Configuration} & \textbf{Value} \\
\midrule
Training pairs          & 1,200,000 \\
Validation pairs        & 50,000 \\
Epochs                  & 6 \\
Batch size              & 1024 \\
Optimizer               & AdamW \\
Learning rate           & \(10^{-4}\) \\
Weight decay            & 0.01 \\
Gradient clipping       & 1.0 \\
Training dtype          & bfloat16 \\
Seed                    & 20260603 \\
Shuffle                 & Yes \\
LR scheduler            & None \\
Relation mode           & \texttt{rich\_pair} \\
Coarse classification   & Hard cross-entropy \\
Ordinal head            & None \\
Base VLM update         & None \\
\bottomrule
\end{tabular}
\end{table}

\paragraph{Internal validation and checkpoint selection.}
The training pipeline uses the same fixed internal validation procedure for all three decoders. Table~\ref{tab:rotationcue_internal_val} reports the selected checkpoint and its hierarchical-state accuracy on the independent 50,000-pair validation set.

Same-object accuracy is measured on all 50,000 pairs. Rotation-present accuracy is evaluated on the 37,500 same-object pairs for which that state is defined, while angle-range accuracy is evaluated on the 32,500 pairs with an actual rotation. Joint hierarchical accuracy requires all active fields of the final state to be correct.

\begin{table}[!ht]
\centering
\small
\setlength{\tabcolsep}{5pt}
\caption{Internal validation performance (\%) of the selected RotationCue decoder checkpoints.}
\label{tab:rotationcue_internal_val}
\begin{tabular}{lccccc}
\toprule
Model & Epoch & Same-object & Rotation-present & Angle-range & Joint \\
\midrule
Qwen3-VL-8B
& 5 & \textbf{98.70} & \textbf{98.68} & 67.44 & \textbf{77.75} \\
Qwen2.5-VL-7B
& 6 & 98.03 & 98.01 & 63.85 & 74.95 \\
InternVL3.5-8B
& 5 & 97.28 & 97.26 & \textbf{67.65} & 76.47 \\
\bottomrule
\end{tabular}
\end{table}

Across the three models, the two gating states achieve 97.26\%--98.70\% accuracy, and angle-range accuracy is 63.85\%--67.65\%.

\subsection{RotationCue for Multi-Image Tasks}
\label{sec:appendix_rotationcue_multiimage}

RotationCue itself remains strictly pairwise. For a multi-image task, we first construct a task-specific set of image pairs
\[
\mathcal{P}
=
\{(I_{i_k},I_{j_k})\}_{k=1}^{K},
\]
and independently predict one rotation state for each pair,
\[
\hat z_k
=
D_\phi\bigl(R_{\mathrm{full}}(F_{i_k},F_{j_k})\bigr),
\qquad
k=1,\ldots,K.
\]

Each serialized pairwise state forms a context card. The cards are concatenated as structured context,
\[
C_{\mathrm{multi}}
=
C(\hat z_1)\oplus\cdots\oplus C(\hat z_K),
\]
and supplied together with all original images to the frozen VLM,
\[
\hat y
=
M_\theta
\bigl(
\mathcal{I},
q,
C_{\mathrm{multi}}
\bigr).
\]

The frozen VLM receives all original images and the pairwise state cards and generates the final answer in a single inference step. No voting or programmatic aggregation is applied to the pairwise predictions.

\begin{table}[!ht]
\centering
\caption{Pair construction for multi-image OR-Bench tasks.}
\label{tab:rotationcue_multiimage}
\small
\setlength{\tabcolsep}{8pt}
\renewcommand{\arraystretch}{1.12}
\begin{tabular}{@{}lcc@{}}
\toprule
\textbf{Task}
& \textbf{Constructed pairs}
& \textbf{\# Cards} \\
\midrule
View Matching
& Ref--A, Ref--B, Ref--C
& 3 \\
Post-Rotation View
& Ref--A, Ref--B, Ref--C, Ref--D
& 4 \\
Rotation Comparison
& A1--A2, B1--B2
& 2 \\
\bottomrule
\end{tabular}
\end{table}

\subsection{Structured State Injection and Controls}
\label{sec:appendix_rotationcue_injection}

At inference time, the decoder predicts
\[
\hat z
=
(\hat z_s,\hat z_r,\hat z_c),
\]
which is serialized using a fixed field order:

\begin{verbatim}
same_object: {yes|no}
rotation_present: {yes|no|not_applicable}
angle_range: {small|medium|large|not_applicable}
\end{verbatim}

If \texttt{same\_object=no}, the remaining fields are set to \texttt{not\_applicable}. If \texttt{same\_object=yes} and \texttt{rotation\_present=no}, only \texttt{angle\_range} is set to \texttt{not\_applicable}. An angle range is provided only when both gates are positive.

The serialized state is inserted before the task question. The frozen VLM uses this structured context together with the original images for subsequent reasoning.

The complete inference path is
\[
(I_A,I_B)
\rightarrow
(F_A,F_B)
\rightarrow
\hat z
\rightarrow
C(\hat z)
\rightarrow
M_\theta
\bigl(
I_A,I_B,q,C(\hat z)
\bigr).
\]

The main-text context-source comparison includes Raw, zero-shot CoT, 5-shot CoT, Self-gen, and RotationCue. Self-gen prompts the base model to generate the same three-field context before producing its final answer, holding the field structure fixed while changing the source of its content.

We use three controlled variants to examine the role of state quality.

\paragraph{Shuffled RotationCue.}
Predicted states are deterministically permuted across evaluation examples,
\[
x_i
\rightarrow
C(\hat z_{\pi(i)}),
\]
breaking state--input correspondence while preserving the same state format and marginal text distribution.

\paragraph{RotationCue.}
Each example receives the state recovered from its own frozen representations,
\[
x_i
\rightarrow
C(\hat z_i).
\]

\paragraph{Oracle RotationCue.}
The same interface is populated with the metadata-derived ground-truth state,
\[
x_i
\rightarrow
C(z_i^\star).
\]

Oracle RotationCue provides a ground-truth-state reference. The aggregate state-quality results are reported in the main text; across all three models they follow
\[
\mathrm{Shuffled}
<
\mathrm{Raw}
<
\mathrm{RotationCue}
<
\mathrm{Oracle}.
\]

Because Shuffled and Oracle retain the same structured interface while altering only state--input correspondence or state correctness, these controls distinguish useful state recovery from a generic benefit of adding rotation-related text.

\subsection{RotationCue Context Field Ablation}
\label{sec:appendix_rotationcue_context_ablation}

We ablate the fields included in the RotationCue context to examine their contributions. We compare the full context with two partial variants: \textit{Same + Rotation}, which retains only the same-object and rotation-presence fields, and \textit{Angle Range}, which retains only the predicted coarse angle range.

\begin{table}[!ht]
\centering
\caption{Ablation of the fields included in the RotationCue context. Values are Macro-8 scores (\%); values in parentheses are gains over Raw ($\Delta$, percentage points).}
\label{tab:rotationcue_context_ablation}
\small
\setlength{\tabcolsep}{5pt}
\renewcommand{\arraystretch}{1.12}
\begin{tabular}{@{}lcccc@{}}
\toprule
\textbf{Base VLM} &
\textbf{Raw} &
\textbf{Same + Rotation} &
\textbf{Angle Range} &
\textbf{Full RotationCue} \\
\midrule
Qwen3-VL-8B
& 46.8
& 48.5 $(+1.7)$
& 56.4 $(+9.6)$
& \textbf{58.0} $(+11.2)$ \\
Qwen2.5-VL-7B
& 42.3
& 43.4 $(+1.1)$
& 50.7 $(+8.4)$
& \textbf{54.9} $(+12.6)$ \\
InternVL3.5-8B
& 43.0
& 45.6 $(+2.6)$
& 50.3 $(+7.3)$
& \textbf{50.9} $(+7.9)$ \\
\bottomrule
\end{tabular}
\end{table}

Across all three VLMs, the angle-range field provides most of the improvement over Raw, whereas the same-object and rotation-presence fields alone yield only modest gains. The full RotationCue context remains strongest, indicating that the additional fields provide complementary information.

\subsection{Efficiency and Stochastic-Decoding Robustness}
\label{sec:appendix_rotationcue_efficiency}

Under the same inference setup, RotationCue processes an example in approximately \(0.9\) seconds on average, compared with \(2.2\) seconds for zero-shot CoT, and adds only \(1.2\%\) latency over Raw. This corresponds to approximately \(2.4\times\) faster inference than zero-shot CoT. Decoder training requires \(0.23\) A100 GPU-hours in the Qwen3-VL-8B comparison reported in the main text.

We further evaluate robustness when the final answer is sampled rather than greedily decoded. On Qwen3-VL-8B, 400 examples are independently decoded nine times using
\[
T=0.5,
\qquad
\operatorname{top\text{-}p}=0.95.
\]

Accuracy standard deviation is the population standard deviation across the nine run-level accuracies. Exact Consistency measures the fraction of examples for which all nine sampled answers are identical,
\[
\operatorname{EC}=\frac{1}{N}\sum_{i=1}^{N}\mathbb{1}\left[\hat y_i^{(1)}=\cdots=\hat y_i^{(9)}\right],
\]
and
\[
\operatorname{FlipRate}=1-\operatorname{EC}.
\]
We additionally report Correct-Stable, the fraction of examples for which all nine sampled answers equal the ground-truth answer,
\[
\operatorname{CS}=\frac{1}{N}\sum_{i=1}^{N}\mathbb{1}\left[\hat y_i^{(1)}=\cdots=\hat y_i^{(9)}=y_i\right].
\]
Prediction entropy summarizes the dispersion of the nine sampled answers for each example; lower values indicate a more concentrated prediction distribution.

\begin{table}[!ht]
\centering
\caption{Accuracy and prediction stability under stochastic decoding on Qwen3-VL-8B. Each of 400 examples is decoded nine times with temperature \(0.5\) and top-\(p=0.95\).}
\label{tab:rotationcue_stochastic}
\scriptsize
\setlength{\tabcolsep}{4pt}
\renewcommand{\arraystretch}{1.12}
\begin{tabular}{@{}lrrrrrr@{}}
\toprule
\textbf{Method} & \textbf{Acc. Mean} & \textbf{Acc. Std} & \textbf{Exact Cons.} & \textbf{Flip Rate} & \textbf{Entropy} & \textbf{Correct-Stable} \\
\midrule
Raw & 56.6 & 0.7 & 84.0 & 16.0 & 0.131 & 50.5 \\
Zero-shot CoT & 58.2 & 1.2 & 71.3 & 28.8 & 0.263 & 47.3 \\
RotationCue & \textbf{64.8} & \textbf{0.5} & \textbf{86.0} & \textbf{14.0} & \textbf{0.111} & \textbf{59.0} \\
\bottomrule
\end{tabular}
\end{table}

RotationCue retains its accuracy advantage under stochastic decoding while producing more consistent predictions than zero-shot CoT. Its Exact Consistency reaches \(86.0\%\), compared with \(71.3\%\) for zero-shot CoT, while Correct-Stable performance increases from \(47.3\%\) to \(59.0\%\).

\section{External Spatial Evaluation}
\label{sec:appendix_external}

We additionally evaluate the RotationCue inference pipeline on VSR-ZS, CV-Bench, and SpinBench to assess whether the structured-state interface preserves performance on broader spatial reasoning benchmarks. We use 1,222 examples from VSR-ZS, 2,638 from CV-Bench, and the 2,739-example SpinBench test split.

\subsection{Input Construction}
\label{sec:appendix_external_input}

RotationCue always operates on image pairs. For the single-image VSR-ZS and CV-Bench tasks, we duplicate the input image and construct
\[
(I_A,I_B)=(I,I),
\]
so that the original visual content remains unchanged while retaining the pairwise RotationCue interface. The resulting structured state is provided together with the original task input to the frozen VLM. This construction evaluates whether introducing the structured-state interface preserves performance on single-image spatial reasoning tasks.

SpinBench contains heterogeneous single- and multi-image tasks. We therefore use a task-aware adapter that constructs pairwise RotationCue inputs when the task defines a meaningful object relation. For reference-to-candidate tasks, the adapter forms
\[
\mathcal P
=
\{
(\mathrm{Ref},C_1),\ldots,(\mathrm{Ref},C_K)
\},
\]
while explicit two-view rotation tasks use their provided image pair. Identity or odd-one-out tasks may use all pairwise combinations,
\[
|\mathcal P|=\binom{N}{2}.
\]

Each pair is processed independently by the same RotationCue decoder,
\[
\hat z_k=D_\phi\bigl(R_{\mathrm{full}}(F_{i_k},F_{j_k})\bigr),
\]
and all resulting cards are supplied together with the complete original image set and question. The frozen VLM generates the final answer in a single inference step.

The angle field represents unsigned rotation magnitude; task-specific interpretation remains with the VLM.

For tasks without a meaningful pairwise rotation relation, we retain the original Raw prompting. This routing is determined solely by the task structure, independently of ground-truth labels, model predictions, or task performance. Under this fixed policy, RotationCue is invoked for 497 examples and Raw inference is retained for the remaining 2,242 examples. We report the full-test score of the resulting task-aware pipeline.

\begin{table}[!ht]
\centering
\caption{External spatial benchmarks and RotationCue input construction.}
\label{tab:external_rotationcue}
\small
\setlength{\tabcolsep}{8pt}
\renewcommand{\arraystretch}{1.12}
\begin{tabular}{@{}lrl@{}}
\toprule
\textbf{Benchmark}
& \textbf{\# Examples}
& \textbf{RotationCue construction} \\
\midrule
VSR-ZS
& 1,222
& Duplicate single image: \(I\)--\(I\) \\
CV-Bench
& 2,638
& Duplicate single image: \(I\)--\(I\) \\
SpinBench
& 2,739
& Task-aware pair construction with Raw fallback \\
\bottomrule
\end{tabular}
\end{table}

\section{Rotation-SFT: Data Construction and Training Details}
\label{sec:appendix_rotation_sft}

Rotation-SFT provides a supervision-aligned alternative to RotationCue. Both methods use the same external multi-view sources, rotation semantics, and angular intervals, but apply the supervision differently. RotationCue trains an external decoder over frozen internal representations, whereas Rotation-SFT directly adapts the language side of Qwen3-VL-8B through LoRA. The vision encoder and multimodal projector remain frozen.

\subsection{External Training Data and Supervision Alignment}
\label{sec:appendix_sft_data}

Rotation-SFT is constructed from the same independent multi-view sources used for RotationCue, namely ALOI-View~\citep{geusebroek2005aloi} and SelectiveMV ModelNet40v1~\citep{alzahrani2024selectivemv,wu2015shapenets}. We retain the same rotation definitions and coarse angular intervals, so that the supervision semantics are aligned between the two methods.

The full SFT training set contains 900,000 examples. A separate class-balanced validation set contains 20,000 examples and is used for checkpoint selection.

OR-Bench examples are not used to construct the SFT training or validation data.

\subsection{Conversion to Four-Way VQA}
\label{sec:appendix_sft_vqa}

RotationCue represents the relation hierarchically through \texttt{same\_object}, \texttt{rotation\_present}, and \texttt{angle\_range}. For Rotation-SFT, same-object pairs are rewritten as two-image, four-way VQA examples.

The training prompt uses the following fixed prefix:

\begin{verbatim}
<image><image>
Image A is shown first and Image B second.
Both images show the same object.
Which option best describes the relative rotation
from Image A to Image B?
Answer with the letter only.
\end{verbatim}

The answer options correspond to four semantic classes. For a pair with unsigned rotation magnitude \(\theta\), the semantic target is
\[
y_{\mathrm{SFT}}(\theta)
=
\begin{cases}
\texttt{no\_rotation},
& \theta=0^\circ, \\[2pt]
\texttt{small},
& 0^\circ<\theta<60^\circ, \\[2pt]
\texttt{medium},
& 60^\circ\leq\theta\leq120^\circ, \\[2pt]
\texttt{large},
& 120^\circ<\theta\leq180^\circ .
\end{cases}
\]

Thus, Rotation-SFT preserves RotationCue's zero-rotation and angular-range semantics for same-object pairs in a four-way VQA target.

\subsection{Answer-Option Construction}
\label{sec:appendix_sft_options}

We avoid a fixed correspondence between semantic classes and answer letters. The correct answer position is cycled deterministically across classes, while the ordering of the remaining choices is determined from a SHA-256 hash of the source sample identifier. Consequently,
\[
y_{\mathrm{semantic}}
\not\equiv
y_{\mathrm{letter}},
\]
preventing the model from solving the training task through a stable class-to-letter shortcut.

\subsection{Language-Side LoRA Adaptation}
\label{sec:appendix_sft_lora}

Rotation-SFT adapts Qwen3-VL-8B using LoRA on the language side. The vision encoder and multimodal projector remain frozen throughout training. The LoRA configuration is
\[
r=8,
\qquad
\alpha=16,
\qquad
p_{\mathrm{drop}}=0.05,
\]
with the target setting \texttt{all}.

Accordingly, visual feature extraction and multimodal projection remain unchanged, while the language-side LoRA parameters receive the rotation VQA supervision. This differs from RotationCue, for which the entire base VLM remains frozen and only the external decoder is trained.

\subsection{Training Configuration}
\label{sec:appendix_sft_training}

The reported full-scale run uses all 900,000 training examples for one epoch. Table~\ref{tab:rotation_sft_training} summarizes the training configuration.

\begin{table}[!ht]
\centering
\caption{Rotation-SFT training configuration.}
\label{tab:rotation_sft_training}
\small
\setlength{\tabcolsep}{9pt}
\renewcommand{\arraystretch}{1.12}
\begin{tabular}{@{}lc@{}}
\toprule
\textbf{Configuration} & \textbf{Value} \\
\midrule
Base model                & Qwen3-VL-8B \\
Training examples         & 900,000 \\
Validation examples       & 20,000 \\
Epochs                    & 1 \\
LoRA rank                 & 8 \\
LoRA alpha                & 16 \\
LoRA dropout              & 0.05 \\
LoRA target setting       & \texttt{all} \\
Per-device batch size     & 4 \\
Gradient accumulation     & 4 \\
Learning rate             & \(10^{-4}\) \\
LR scheduler              & Cosine \\
Warmup ratio              & 3\% \\
Training dtype            & bfloat16 \\
Vision encoder            & Frozen \\
Multimodal projector      & Frozen \\
Distributed processes     & 2 \\
Hardware                  & \(2\times\) A100 \\
Training time             & \(\sim21\) hours \\
Training cost             & \(\sim42\) A100 GPU-hours \\
\bottomrule
\end{tabular}
\end{table}

\subsection{External Validation and Checkpoint Selection}
\label{sec:appendix_sft_selection}

Checkpoint selection is performed exclusively on the independent 20,000-example class-balanced validation set from the external multi-view training sources. For a candidate checkpoint \(c\), we compute four-way VQA accuracy,
\[
\operatorname{Acc}_{\mathrm{val}}(c)
=
\frac{1}{N_{\mathrm{val}}}
\sum_{i=1}^{N_{\mathrm{val}}}
\mathbb{1}
\left[
\hat y_i^{(c)}=y_i
\right],
\qquad
N_{\mathrm{val}}=20{,}000.
\]

The reported checkpoint is fixed as
\[
c^\star
=
\arg\max_c
\operatorname{Acc}_{\mathrm{val}}(c).
\]

OR-Bench is not used for optimization or checkpoint selection and is evaluated only after \(c^\star\) has been fixed. The resulting protocol is
\[
\text{external training}
\rightarrow
\text{external validation}
\rightarrow
\text{checkpoint selection}
\rightarrow
\text{OR-Bench evaluation}.
\]

\subsection{Task-Level Results}
\label{sec:appendix_sft_results}

For completeness, Table~\ref{tab:rotation_sft_results} reports the complete OR-Bench task-level results of the selected Rotation-SFT checkpoint. All deltas are computed from displayed scores and reported in percentage points.

\begin{table}[!ht]
\centering
\caption{Task-level accuracy (\%) of Qwen3-VL-8B before and after Rotation-SFT. Macro-8 gives equal weight to the eight tasks, whereas Micro-8 aggregates all 2,109 examples. Deltas are computed from displayed scores.}
\label{tab:rotation_sft_results}
\small
\setlength{\tabcolsep}{7pt}
\renewcommand{\arraystretch}{1.12}
\begin{tabular}{@{}lrccc@{}}
\toprule
\textbf{Task} & \textbf{N} & \textbf{Raw} & \textbf{Rotation-SFT} & \(\boldsymbol{\Delta}\) \\
\midrule
Rotation Detection & 486 & \textbf{99.6} & 89.9 & \(-9.7\) \\
Coarse Angle & 243 & 40.7 & \textbf{85.2} & \(+44.5\) \\
\(90^\circ\) Threshold & 243 & 37.0 & \textbf{41.6} & \(+4.6\) \\
Exact Angle (MCQ) & 243 & 24.7 & \textbf{40.7} & \(+16.0\) \\
Exact Angle (Open) & 243 & 25.5 & \textbf{27.6} & \(+2.1\) \\
\midrule
View Matching & 216 & \textbf{65.3} & 54.2 & \(-11.1\) \\
Rotation Comparison & 243 & 48.6 & \textbf{49.8} & \(+1.2\) \\
Post-Rotation View & 192 & \textbf{32.8} & 20.3 & \(-12.5\) \\
\midrule
\textbf{Macro-8} & --- & 46.8 & \textbf{51.2} & \(+4.4\) \\
\textbf{Micro-8} & 2,109 & 53.0 & \textbf{56.3} & \(+3.3\) \\
\bottomrule
\end{tabular}
\end{table}

\end{document}

%% file: math_commands.tex
\usepackage{amsmath,amsfonts,bm}

\def\eqref#1{equation~\ref{#1}}
\def\1{\bm{1}}

\DeclareMathAlphabet{\mathsfit}{\encodingdefault}{\sfdefault}{m}{sl}
\SetMathAlphabet{\mathsfit}{bold}{\encodingdefault}{\sfdefault}{bx}{n}